\documentclass[journal]{IEEEtran}

\ifCLASSINFOpdf

\else

\fi

\usepackage{times}

\usepackage{soul}
\usepackage{url}
\usepackage{subfigure}
\usepackage[utf8]{inputenc}
\usepackage[small]{caption}
\usepackage{graphicx}
\usepackage{amsmath}
\usepackage{booktabs}
\usepackage{algorithm} 
\usepackage{algpseudocode} 
\usepackage{pifont}
\usepackage{xcolor}

\usepackage{makecell,booktabs}
\definecolor{Gray}{gray}{0.95}
\usepackage{colortbl}
\usepackage{tabularx}
\usepackage{amssymb}
\usepackage[numbers,compress]{natbib}
\usepackage{ulem}
\usepackage{color}
\usepackage{bm}

\begin{document}

\title{Reinforcement Learning-based Dialogue Guided Event Extraction to Exploit Argument Relations}

\author{Qian Li,
        Hao Peng,
        Jianxin Li,~\IEEEmembership{~Member,~IEEE},
        Jia Wu,~\IEEEmembership{Senior Member,~IEEE},
        Yuanxing Ning,
        Lihong Wang,\\
        Philip S. Yu,~\IEEEmembership{~Fellow,~IEEE},
        Zheng Wang,~\IEEEmembership{~Member,~IEEE}
\thanks{Manuscript received May 2021, revised September 2021, accepted December 2021. This work was supported by the NSFC through grants (No.U20B2053, and 62002007), State Key Laboratory of Software Development Environment (SKLSDE-2020ZX-12), NSF (III-1526499, III-1763325, III-1909323), and the UK EPSRC (EP/T01461X/1). This work was also sponsored by CAAI-Huawei MindSpore Open Fund. Thanks for computing infrastructure provided by Huawei MindSpore platform. (\emph{Corresponding author: Jianxin Li.})}
\IEEEcompsocitemizethanks{\IEEEcompsocthanksitem Qian Li, Hao Peng, Jianxin Li and Yuanxing Ning are with Beijing Advanced Innovation Center for Big Data and Brain Computing, Beihang University, Beijing 100191, China. E-mail: \{liqian, penghao, lijx, ningyx\}@act.buaa.edu.cn.
\IEEEcompsocthanksitem Jia Wu is with the Department of Computing, Macquarie University, Sydney, Australia. E-mail: jia.wu@mq.edu.au.
\IEEEcompsocthanksitem Lihong Wang is with the National Computer Network Emergency Response Technical Team/Coordination Center of China, Beijing 100029, China. Email: wlh@isc.org.cn. 
\IEEEcompsocthanksitem Philip S. Yu is with the Department of Computer Science, University of Illinois at Chicago, Chicago 60607, USA. E-mail: psyu@uic.edu.
\IEEEcompsocthanksitem Zheng Wang is with the School of Computing, University of Leeds, Leeds LS2 9JT, UK. E-mail: z.wang5@leeds.ac.uk.}
}

\markboth{IEEE/ACM Transactions on Audio, Speech and Language Processing,~Vol.~21, No.~12, May~2021}%
{Shell \MakeLowercase{\textit{et al.}}: Bare Demo of IEEEtran.cls for IEEE Journals}

\maketitle

\begin{abstract}
Event extraction is a fundamental task for natural language processing. Finding the roles of event arguments like event participants is essential for event extraction. However, doing so for real-life event descriptions is challenging because an argument’s role often varies in different contexts. While the relationship and interactions between multiple arguments are useful for settling the argument roles, such information is largely ignored by existing approaches. This paper presents a better approach for event extraction by explicitly utilizing the relationships of event arguments. We achieve this through a carefully designed task-oriented dialogue system. To model the argument relation, we employ reinforcement learning and incremental learning to extract multiple arguments via a multi-turned, iterative process. Our approach leverages knowledge of the already extracted arguments of the same sentence to determine the role of arguments that would be difficult to decide individually. It then uses the newly obtained information to improve the decisions of previously extracted arguments. This two-way feedback process allows us to exploit the argument relations to effectively settle argument roles, leading to better sentence understanding and event extraction. Experimental results show that our approach consistently outperforms seven state-of-the-art event extraction methods for the classification of events and argument role and argument identification. 
\end{abstract}


\IEEEpeerreviewmaketitle

\section{Introduction}\label{sec:Introduction}

\IEEEPARstart{E}{vent} extraction aims to detect, from the text, the occurrence of events of specific types and to extract arguments (e.g., typed event participants or other attributes) that are associated with an event \cite{DBLP:journals/cacm/CowieL96}. It is a fundamental technique underpinning many Natural Language Processing (NLP) tasks like knowledge reasoning \cite{DBLP:books/daglib/0040913}, text summarization \cite{DBLP:journals/ir/Liddy01}, and event prediction \cite{DBLP:reference/db/Wasserkrug09}.

\begin{figure}[!t]
 \centering
 \includegraphics[width=\linewidth]{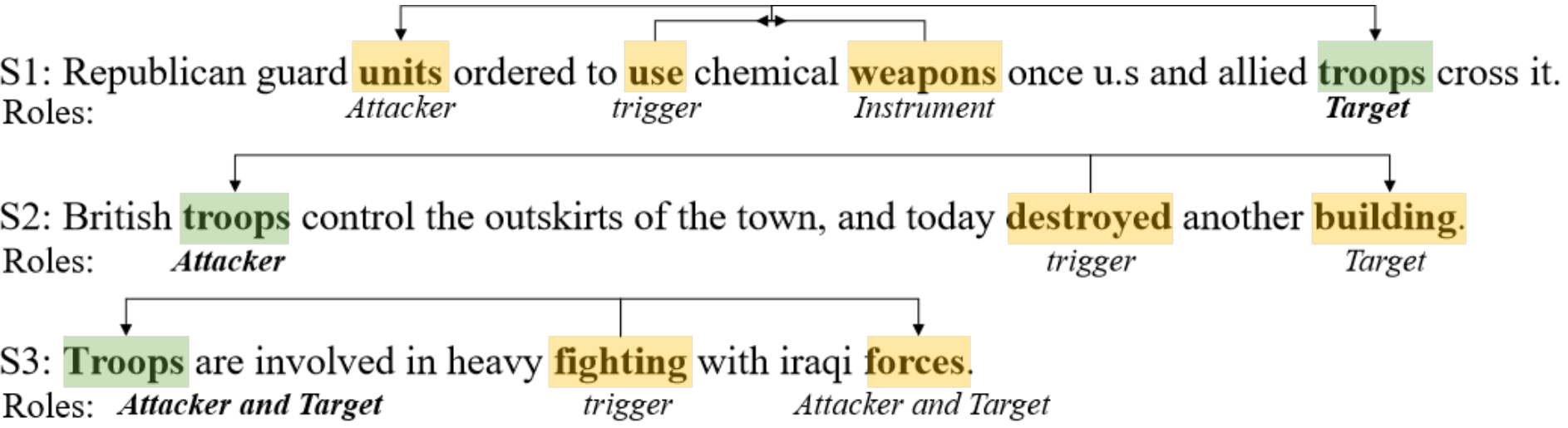}
 \caption{Three example sentences belong to the ``Conflict: Attack" event type from the ACE 2005 dataset.}
 \label{exampleACE}
 \vspace{-3mm}
\end{figure}

Event extraction requires extracting all arguments and their roles corresponding to each event. Doing so is challenging because an event is often associated with more than one argument, whose role can vary in different contexts. For example, the argument ``troops" has different roles in multiple sentences, as shown in Fig.~\ref{exampleACE}. This argument has the role of being the ``\emph{target}" in sentence $S1$, while in sentence $S2$, its role is the ``\emph{attacker}". In sentence $S3$, the argument ``troops" could be either the ``\emph{target}" or the ``\emph{attacker}". 
To extract an event, we need to identify the argument role correctly. Failing to do so can lead to erroneous information propagation, affecting the recognition of other event arguments and sentence understanding. For example, incorrectly associating the ``troops” argument in the sentence $S2$ as the \emph{target} will lead to misunderstanding of the sentence. Unfortunately, argument role detection remains an open problem because an argument can be associated with multiple roles.

\begin{figure*}[!t]
    \centering
    \includegraphics[width=\linewidth]{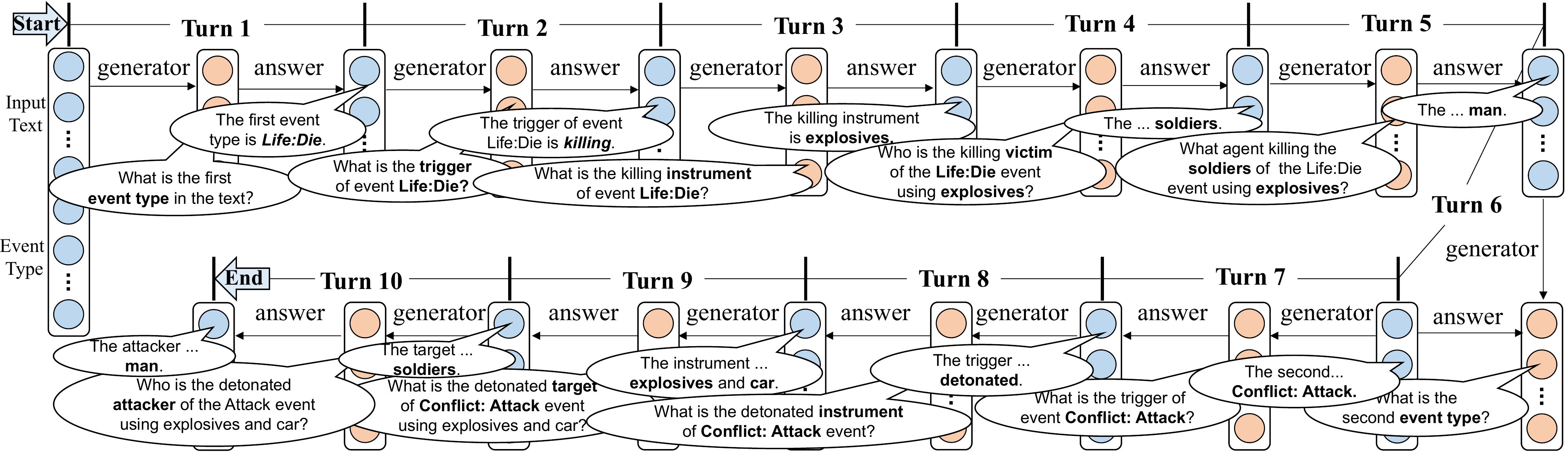}
    \caption{An example of our dialogue guided event extraction. The input sentence is: \uline{``As the soldiers approached, the man detonated explosives in the car killing all four of the soldiers".} It needs 10 turns to complete event extraction on the sentence. } 
    \label{dialogue}
\end{figure*}

Our work aims to find new ways to identify event argument roles for event extraction. \textbf{Our key insight is that multiple arguments associated with an event are typically strongly correlated}. Their correlations can provide useful information for determining the role of an event argument. Consider again our example given earlier in Fig.~\ref{exampleACE}. To determine the role of argument ``troops" in the sentence $S1$, we can consider its relevant arguments of ``weapons" and ``use”. The roles and appearance order of ``weapons" and ``use” suggest ``troops" is the target in the context. Although the argument ``use” has many subtle roles, we can still attribute it with the \emph{``Conflict: Attack"} event type in sentence $S1$ by using ``weapons" as a hint. Moreover, the role of argument ``troops" can be recognized to role by looking at the argument ``weapons" and ``use". If the ``troops" is detected firstly, it may misidentify its argument role and other arguments. As can be seen from this representative example, the relation among arguments can help in inferring the argument roles that are essential for event extraction. However, prior work largely ignores such relations, leaving much room for improvement. 
Most existing methods extract all arguments simultaneously \cite{DBLP:conf/aaai/ShaQCS18, DBLP:conf/emnlp/LiuLH18} or individual arguments sequentially \cite{DBLP:conf/emnlp/DuC20}, all of which do not consider the effect of argument extraction order.
Our work seeks to close this gap by explicitly modeling the argument relation for event extraction.

Our approach is enabled by the recent advance in task-oriented dialogue systems \cite{DBLP:conf/aaai/LiptonLG00D18, DBLP:conf/acl/ZhangLGC19} that are shown to be highly effective in entity-relation extraction \cite{DBLP:conf/acl/LiYSLYCZL19}.
A task-oriented dialogue system uses domain knowledge, e.g., knowledge structures of intentions extracted from sentences, to complete a specific task. The natural language understanding task is structured as many slots are to be filled, where each slot can take a set of possible values. For example, a travel query could be translated into a structure consisting of slots like \emph{original\_city}, \emph{destination\_city}, \emph{departure\_time} and \emph{arrival\_time}. The goal of the dialogue system in this context is thus to extract the right values from the user sentence to fill the slots. The recent progress in a task-oriented dialogue system allows one to effectively exploit the dialogue historical information to optimize slot filling with the right order \cite{DBLP:conf/acl/MrksicSWTY17, DBLP:conf/emnlp/ZhangSKWSLLZS20, DBLP:conf/aaai/Sun0XYX20}.

In this work, we formulate the event extraction task within a task-oriented dialogue system, as shown in Fig.~\ref{dialogue}. We consider the problem of event extraction as filling slots of relevant arguments and their roles extracted from the input sentence. To this end, we develop a multi-turn dialogue system \cite{DBLP:conf/acl/ZhangLGC19} with two agents to iteratively solve the slot filling problem. During each turn, one agent selects an argument role and generates a query through a dialogue generator. For example, the query for the role of ``instrument" could be ``What is the killing instrument of event \emph{Life: Die}?" in Turn 3 of Fig.~\ref{dialogue}. The other agent answers the query by identifying the right argument or event type from the sentence. This iterative generation and answering paradigm enable us to introduce knowledge obtained from previous turns when extracting a current argument. For instance, it can exploit the argument relation like ``weapon”, ``use" and ``troops" in sentence $S1$ in Fig.~\ref{exampleACE} to improve the quality of event extraction. 
Our dialogue-based method extracts the argument of victim \emph{"soldiers"} according to the argument of trigger \emph{killing} and instrument \emph{explosives} for event \emph{Life: Die} in Fig.~\ref{dialogue}.
This multi-turn process also enables us to leverage additional information about the newly extracted argument to update and correct the argument roles identified in the previous turn. As we repeat the process, we will obtain more information about event arguments and better understand the sentence over time. This richer information helps us extract argument roles more precisely towards the end of the process.

While our multi-turn dialogue system provides a potentially powerful event extraction capability, its potential can only be fully unlocked if the arguments are processed in the right order. Since we extract arguments and determine their roles in sequential order by utilizing the knowledge obtained from previously extracted arguments, the order of argument extraction is crucial. Ideally, we would like to start from event arguments whose argument roles are likely to be accurately decided using already extracted information and leave the more challenging ones later once we have obtained sufficient information from others. For instance, we may wish to extract argument ``weapon" before ``use" of sentence $S1$ given in Fig.~\ref{exampleACE} because determining the former's role is more straightforward than doing that for the latter. 

\textbf{We address the challenge of argument extraction order by employing Reinforcement Learning (RL) to rank arguments to best utilize the argument relation}. To allow RL to navigate the potentially large problem space, we need to find the right representation of each word in the target sentence and use the representation to predict the start and end position of each argument. 
To that end, \textbf{we use both a lexicon-based graph attention network \cite{DBLP:conf/emnlp/GuiZZPFWH19} and an event-based BERT model \cite{DBLP:conf/nips/VaswaniSPUJGKP17} for learning the word representation from semantics and context two perspectives}. 
We then utilize the learned representation to determine which argument to extract and in what order. We go further by designing an incremental learning strategy to iteratively incorporate the argument relation into the multi-turned event extraction process by continually updating the event representation across turns. By doing so, the representation becomes increasingly more accurate as the argument extraction process proceeds, which, in turn, enhances the quality of the resulting argument and event extraction.

We evaluate our approach\footnote{Code and data are available at: \url{https://github.com/xiaoqian19940510/TASLP-EAREE}} by applying it to sentence-level event extraction performed on the ACE 2005 dataset \cite{DBLP:conf/lrec/DoddingtonMPRSW04}. We compare our approach to 7 recently proposed event extraction approaches \cite{DBLP:conf/aaai/ShaQCS18,DBLP:conf/emnlp/LiuLH18,DBLP:conf/aaai/NguyenN19,DBLP:journals/dint/ZhangJS19,DBLP:conf/acl/YangFQKL19,DBLP:conf/emnlp/DuC20,DBLP:conf/emnlp/LiPCWPLZ20}. Experimental results show that our approach can effectively utilize the argument relation to identify the argument roles, leading to better event extraction performance. We show that our incremental event learning strategy is particularly useful when the amount of labeled data is limited.

This paper makes the following contributions. It is the first to:
\begin{itemize}
\itemsep0em 
\item develop a multi-turned, task-oriented dialogue guided event extraction framework aiming to fill arguments extracted from input text for specific arguments roles (Section \ref{sec:Framework}); 
\item employ reinforcement learning to rank argument extraction order to utilize argument correlation for event extraction (Section \ref{sec:ArgumentRepresentation});
\item leverage a lexicon-based graph attention network and event-based BERT, under an incremental learning framework to learn word representation for even extraction (Sections \ref{sec:EventRepresentation});
\end{itemize}

\section{Related Work}\label{sec:RelatedWork}
  \vspace{-1mm}

Event extraction \cite{DBLP:conf/acl/HuangCFJVHS16, DBLP:conf/naacl/YangM16, DBLP:conf/naacl/FergusonLWH18} is a form of information representation to extract what users are interested in massive data and present it to users in a certain way.
For event extraction task, it can be divided into four subtasks: event classification, trigger identification, argument identification, and argument role classification.
Most recent event extraction works are based on a neural network architecture like Convolutional Neural Network (CNN) \cite{DBLP:conf/acl/ChenXLZ015, DBLP:conf/nlpcc/ZhangXC16}, Recurrent Neural Network (RNN) 
\cite{DBLP:conf/emnlp/NguyenG16, DBLP:conf/aaai/ShaQCS18}, Graph Neural Network (GNN) \cite{DBLP:conf/emnlp/LiuLH18,DBLP:conf/emnlp/HuangYP20}, Transformer \cite{DBLP:conf/acl/YangFQKL19, DBLP:conf/emnlp/LiuCLBL20}, or other networks \cite{DBLP:conf/ijcai/ZhangQZLJ19, DBLP:conf/coling/HuangZTTX20}. 
The method of event extraction based on deep learning first adopts pipeline. The pipeline-based method \cite{DBLP:conf/acl/ChenXLZ015, DBLP:conf/acl/YangFQKL19, DBLP:conf/aaai/ZengFMWYSZ18} is the earliest event extraction based on neural networks and extracts event arguments by utilizing triggers.
It realizes event trigger identification, event classification, event argument identification, and argument role classification tasks successively \cite{DBLP:journals/access/XiangW19} and takes the results of previous tasks as prior knowledge. 
The first two tasks are usually called event detection and the last two tasks are called argument extraction. 
Chen et al. \cite{DBLP:conf/acl/ChenXLZ015} and Nguyen et al. \cite{DBLP:conf/emnlp/NguyenG16, DBLP:conf/acl/NguyenG15} use the CNN model to capture sentence-level clues and overcome complex feature engineering compared with traditional feature-based approaches.
DBRNN \cite{DBLP:conf/aaai/ShaQCS18} have been proved to be influential in introducing graph information into event extraction tasks. 
JMEE \cite{DBLP:conf/emnlp/LiuLH18} is proposed with an attention-based GCN, learning syntactic contextual node representations through first-order neighbors of the graph.
As we all know, an argument usually plays different roles in different events, enhancing the difficulty of the event extraction task.
Yang et al. \cite{DBLP:conf/acl/YangFQKL19} propose a pre-trained language model-based event extractor \cite{DBLP:conf/naacl/DevlinCLT19} to learn contextualized representations proven helpful for event extraction. It separates argument predictions according to the roles and overcomes the argument overlap problem.
However, this requires a high accuracy of trigger identification.
A wrong trigger will seriously affect the accuracy of argument identification and argument role classification.
Therefore, the pipeline based method considers the event trigger as the core of an event \cite{DBLP:conf/aaai/ShaQCS18, DBLP:conf/lpkm/MejriA17, DBLP:conf/cncl/ChenLHL016} and requires high accuracy of event trigger identification, avoiding an adverse effect on event argument extraction. 
In order to overcome the propagation of error information caused by event detection, joint-based event extraction methods are proposed.
The joint event extraction method avoids the influence of trigger identification error on event argument extraction.
It reduces the propagation of error information by combining trigger identification and argument extraction tasks.

The existing event extraction corpus has a few labeled data and hard to expand, such as ACE 2005 with only 599 annotated documents \cite{DBLP:conf/lrec/DoddingtonMPRSW04}.
Existing deep learning-based approaches usually require lots of manually annotated training data. Consequently, except for the difficulty of event extraction itself, inadequate training data also hinders. 
The zero-shot learning method is the right choice, which has been widely applied in NLP tasks \cite{DBLP:conf/icml/Romera-ParedesT15}.
Based on this, Huang et al. \cite{DBLP:conf/acl/DaganJVHCR18} design a transferable neural architecture and stipulate a graph structure to transfer knowledge from the existing types to the extraction of unseen types. 
It finds the event types graph structure, which learns representation almost matching representations from the parsed AMR structure \cite{DBLP:conf/acllaw/BanarescuBCGGHK13}.
Existing event extraction systems, which usually adopt a supervised learning paradigm, have to rely on labelled training data, but the scarcity of high-quality training data is a common problem \cite{DBLP:conf/aclevents/MitamuraYHSBKS15,DBLP:conf/aaai/ZengFMWYSZ18}.

Machine Reading Comprehension (MRC) tasks extract a span from text \cite{DBLP:conf/ijcai/ZhaoYCL20}, is a basic task of question answering.
MRC based event extraction \cite{DBLP:conf/emnlp/LiuCLBL20, DBLP:conf/emnlp/DuC20} designs questions for each argument, helps capture semantic relationship between question and input sentence.
Question answering methods are emerging as a new way for extracting important keywords from a sentence \cite{DBLP:conf/emnlp/LiuCLBL20, DBLP:conf/emnlp/DuC20}. By incorporating domain-knowledge to the question set, one can guide the extraction framework to focus on essential semantics to be extracted from a sentence. 
Existing approaches do not utilize the relations among multiple arguments, leaving much room for improvement. 
Our work aims to close this gap by using the argument relations to infer roles of arguments that are hard to settle in isolation, leading to better performance for argument and event classification.

\begin{figure*}[!t]
    \centering
    \includegraphics[width=\linewidth]{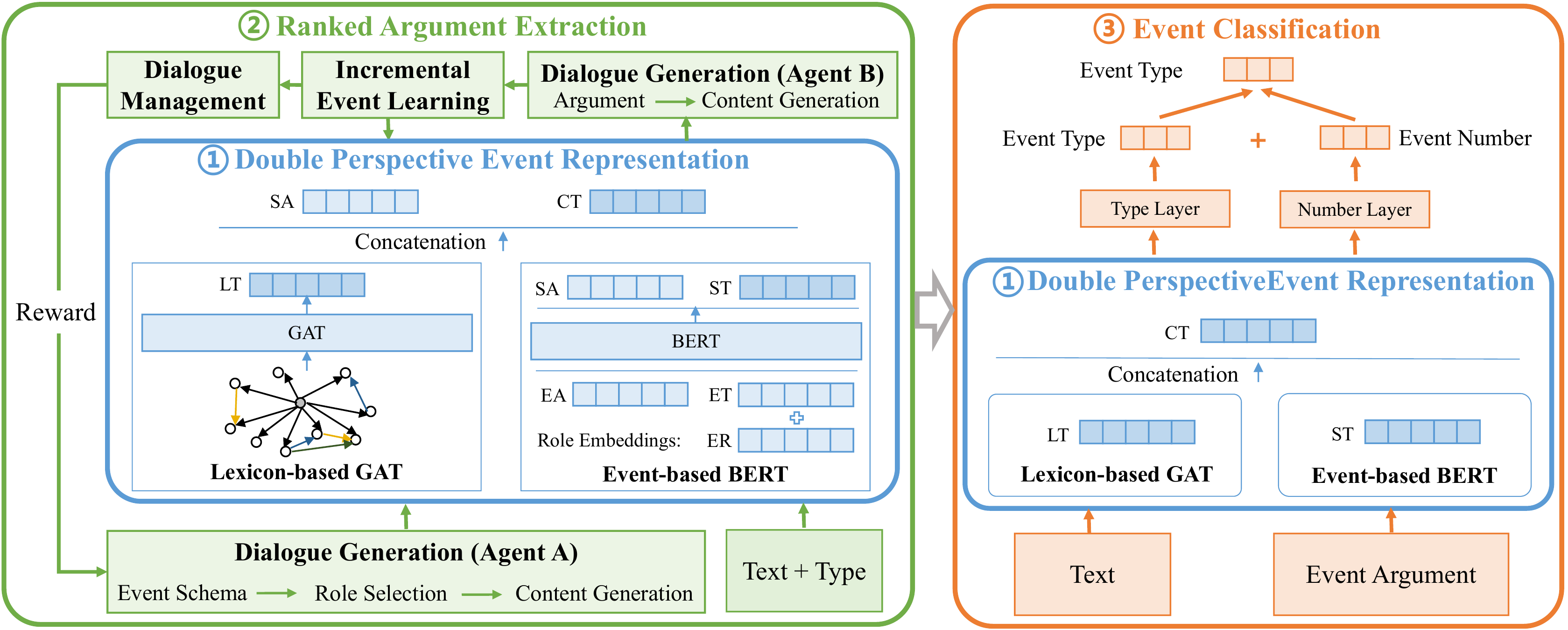}
    \caption{Overview of our event extraction framework. The training flow is: \ding{173} $\rightarrow$ \ding{174}; the testing flow during deployment is: \ding{174} $\rightarrow$ \ding{173} $\rightarrow$ \ding{174}. $LT$ is the lexicon-based word representation. $EA$, $ER$, and $ET$ are the dialogue content embedding of \emph{Agent A}, argument role embedding, and event type embedding. $SA$ and $ST$ are the dialogue content representation of \emph{Agent A} and representation of $ET$ after BERT. $CT$ is the final representation of input text. } 
    \label{Framework}
\end{figure*}

Task-oriented dialogue systems aim to assist the user to complete specific tasks according to existing corpora \cite{DBLP:conf/acl/ZhangLGC19, DBLP:conf/ijcai/MiHZF19}.
The typical task-oriented dialogue system has four components: natural language understanding, dialogue state tracker, dialogue policy learning, and natural language generation \cite{DBLP:journals/sigkdd/ChenLYT17}.
Ramadan et al. \cite{DBLP:conf/acl/RamadanBG18} introduce an approach utilizing semantic correlation in ontology terminologies and dialogue utterances.
It easily utilizes history knowledge for current dialogue \cite{DBLP:conf/aaai/LiptonLG00D18, DBLP:conf/acl/ZhangLGC19}, aiming to build the connection among dialogue.
Therefore, there have been some explorations on formulating NLP tasks as a dialogue task.
It can overcome the inadequacy problem of historical information utilization in MRC.
Therefore, there have been some explorations on formulating NLP tasks as a dialogue task.
We treat event extraction as a dialogue for capturing relationships among arguments. 
Our work leverages the recent development in task-oriented dialogue systems \cite{DBLP:conf/acl/ZhangLGC19, DBLP:conf/ijcai/MiHZF19}. Such a system decouples the problem to be solved through a multi-turn dialogue, allowing the system to connect and exploit information obtained during multiple conversations to solve a new task \cite{DBLP:conf/aaai/LiptonLG00D18, DBLP:conf/acl/ZhangLGC19}.

\section{Event Extraction Framework}\label{sec:Framework}

Fig.~\ref{Framework} gives an overview of our framework that consists of three components for \textbf{\emph{(1) double perspective event representation, (2) ranked argument extraction, and (3) event classification}}. The argument extraction module automatically generates the dialogue based on the event type and the selected predicted arguments.
The selected arguments are produced by incremental event learning method to add the pseudo label as the training data and append the pseudo relation to the lexicon-based graph. The pseudo label is the arguments predicted by our ranked argument extraction model. The pseudo relation is the argument role of the predicted argument. 

We add a pseudo edge by appending an edge among predicted arguments of an event type to update word representation using existing prediction results.
It takes the sentence and event type as input.
Event classification module detects whether the input sentence is an event and classifies the event type to which the text belongs. We design a multi-task learning module to calculate the combined loss of the two tasks to overcome the event type imbalance results in a low recall.
For different event type, different event schema are designed for extracting different arguments according to the schema. 
Our framework is first trained offline using a small amount of labeled data. 
To expand the training data, we design a dialogue generation module, generating multiple question-answering pairs for each trigger or argument for data enhance. 
The trained models can then be applied to extract event types and associated event arguments. 

During the training phase, the reinforcement learning-based, dialogue-guided argument extraction model learns how to extract event arguments by taking as input the target sentence and a label of the event type. 
Our framework will first learn several rounds of conversational argument extraction according to event types and sentences, and train the event classification model according to event arguments. 
In each turn, the predicted argument is provided as a pseudo relation in the lexicon-based graph and a pseudo label in role embeddings of event-based BERT used by the incremental event learning method.
It updates the textual representation by adding the pseudo argument knowledge.
The event classification model is then trained to predict the event type using the pseudo relation knowledge provided by the event extraction module.

During deployment, we used the trained models in an iterative process to perform event extraction. 
We will first predict the event type with the event classification model, and then implement argument extraction according to the predicted event type. So the model ends up running through all the predicted event types.
To do so, we first use the event classification model to predict the event type without a pseudo label and relation. Next, we use the argument extraction model to identify all arguments associated with the predicted event type. We then go back to ask the event classification module again to update the event type using the pseudo label and argument relations extracted by the argument extraction model. This 2-stage iterative process uses the predicted event type to extract arguments, and the extracted information helps the event classification model improve its prediction.

\begin{figure*}[!t]
    \centering
    \includegraphics[width=\linewidth]{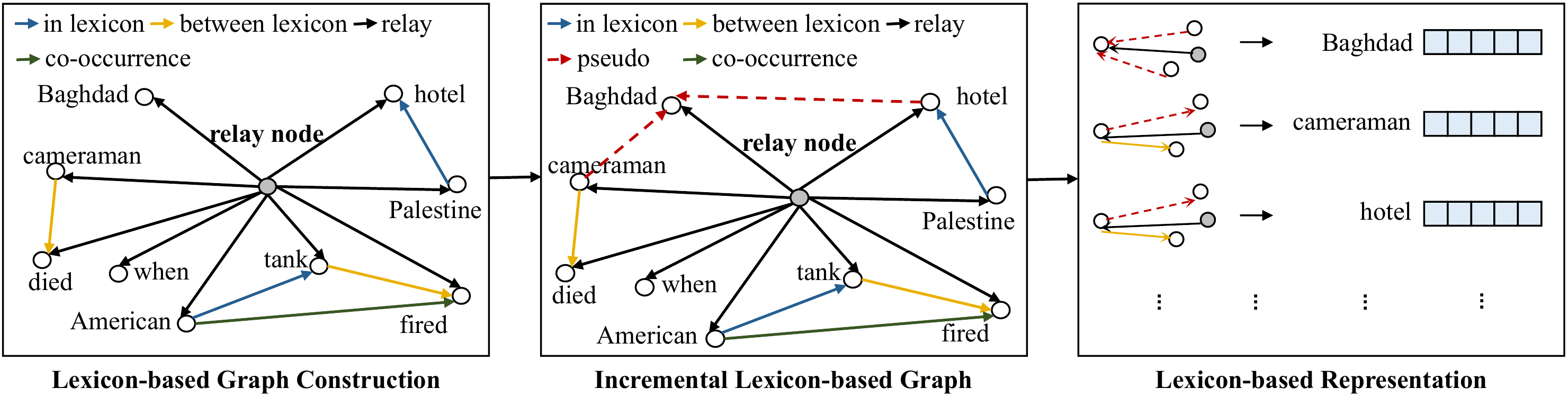}
    \caption{The flowchart of lexicon-based representation. The input sentence is ``In Baghdad, a cameraman died when an American tank fired on the Palestine hotel.". Before constructing the graph, we remove punctuation marks and prepositions from the sentence, which can reduce the complexity of the graph calculation. The graph construction module has five edges: the black edge connects the relay node and all words; the orange edge is the phase's connection; the blue is the association between phrases; the green connects words with high co-occurrence probability; the red is the pseudo argument relation.}
    \label{graph}
\end{figure*}

\subsection{Double Perspective Event Representation}
\label{sec:EventRepresentation}

The first step of our argument extraction pipeline is to learn the representation (or embeddings) to be used for argument selection. We do so by first constructing a lexicon-based graph, from which we learn the lexicon-based representation of individual words. We then learn the context representation at the sentence-level across multiple words. 
To that end, we use both a lexicon-based graph attention network \cite{DBLP:conf/emnlp/GuiZZPFWH19} and an event-based BERT model \cite{DBLP:conf/nips/VaswaniSPUJGKP17} for learning the word representation from semantics and context two perspectives. 

\subsubsection{\textbf{Lexicon-based Representation}} 

Lexicon-based graph neural network has been designed for the node classification task \cite{DBLP:conf/emnlp/GuiZZPFWH19}, proved to be an effective way to learn global semantics.
Thus, we use lexical knowledge to concatenate characters capturing the local composition and a global relay node to capture long-range dependency.

We convert the sentence to a directed graph (as shown in Fig.~\ref{graph}) where each word is represented as a graph node, and a graph edge represents one of the five relations: words in a lexicon; lexicon to lexicon; a relay node connecting to all nodes; co-occurrence words; and pseudo relation among arguments of an event. 
The first connects words in the phrase sequentially until the last word.
The second is to create a line between phrases that the last word of the current phrase is connected with the latter phrase, and each edge represents the potential characteristics of the word that may exist.
We also use a relay node as a virtual hub, which is connected to all other nodes. 
It gathers all the edges and nodes' information, eliminating the boundary ambiguity between words, and learning long-range dependency.
Therefore, the representation of the relay node can be regarded as the representation of the sentence.
The fourth edge represents pseudo argument relation by connecting the predicted arguments in an event.
The last one is calculating the co-occurrence probability of words within sliding windows in the corpus.
The edge weights are measured by pointwise mutual information (PMI):
\begin{equation}
\operatorname{PMI}\left(w_{i}, w_{j}\right)=\log \frac{p(w_{i}, w_{j})}{p(w_{i}) p(w_{j})}=\log \frac{N_{w_{i}, w_{j}} N_{s}}{N_{w_{i}} N_{w_{j}}},   \label{eq:r1}
\end{equation}
where $N_{w_{i}}, N_{w_{j}}, N_{w_{i}, w_{j}}$ are the number of sliding windows containing word $w_{i}$, $w_{j}$ and both $w_{i}$, $w_{j}$, and $i, j \in [1,N]$. $N_{s}$ is the total number of sliding windows in the corpus.

To learn the word-level representation, we extend the lexicon-based graph attention network (LGAT) that is designed for learning global semantics for node classification \cite{DBLP:conf/emnlp/GuiZZPFWH19}. We extend the LGAT by adding a pseudo edge, i.e., by appending an edge among predicted arguments of an event type. Our goal is to update word representation using existing prediction results.
Given an N-word text $T=\{{T_{1}, T_{2}, \dots, T_{N}}\}$, the model embeds the initial input text $ET=\{{ET_{1}, ET_{2}, \dots, ET_{N}}\}$ through the pre-trained embedding matrix. The embedding of predicted event role $R=\{R_{1}, R_{2}, \dots, R_{N}\}$ is represented as $ER=\{ER_{1}, ER_{2}, \dots, ER_{N}\}$. 
The model takes as input $EI=ET \oplus ER$ where $\oplus$ means concatenation and produce a hidden representation $H^{n}=\{{h^{n}_{1}, h^{n}_{2}, \dots, h^{n}_{N}}\}$, for text $T$.
Here, $h^{n}_{i}$ represents the text feature of the $i$-th word on the $n$-th hidden layer.
Therefore, the final node representation is: 
\begin{equation}
{LT}_{i}=f(H^{N}_{i}), i=0, 1, 2, \dots, N, \label{eq:r2}
\end{equation}
where ${LT}_{0}$ is the relay node used to predict the event type.
It is then used to obtain the optimal decision through the conditional random field (CRF).
The probability of a label sequence $\hat{y}=\hat{c_{1}}, \hat{c_{2}}, \dots, \hat{c_{k}}$ can be defined as follows:
\begin{equation}
p(\hat{y} \mid s)=\frac{\exp \left(\sum_{i=1}^{N} \phi\left(\hat{c}_{i-1}, \hat{c}_{i}, \mathbf{LT}\right)\right)}{\sum_{y^{\prime} \in \mathrm{L}(s)} \exp \left(\sum_{i=1}^{N} \phi\left(c_{i-1}^{\prime}, c_{i}^{\prime}, \mathbf{LT}\right)\right)},
\end{equation}
where $L(s)$ is the set of all arbitrary label sequences.
\begin{equation}
\phi\left(\hat{c}_{i-1}, \hat{c}_{i}, \mathbf{h}\right)=\mathbf{W}_{\left(c_{i-1}, c_{i}\right)} \mathbf{c}_{i}^{T}+\mathbf{b}_{\left(c_{i-1}, c_{i}\right)},
\end{equation}
where $\mathbf{W}_{\left(c_{i-1}, c_{i}\right)}\in \mathbb{R}^{k \times N}$ and $\mathbf{b}_{\left(c_{i-1}, c_{i}\right)}\in \mathbb{R}^{k \times N}$ are the weight and bias parameters specific to the labels $c_{i-1}$ and $c_{i}$, $k$ is the number of event types, and $N$ is the input length of text.

\subsubsection{\textbf{Event-based Context Representation}} 
BERT is a multi-layer bidirectional Transformer \cite{DBLP:conf/nips/VaswaniSPUJGKP17}, achieving significant performance improvement on event extraction task \cite{DBLP:conf/acl/YangFQKL19}.
We use a BERT model \cite{DBLP:conf/acl/YangFQKL19} to learn the context representation. 
Specifically, we feed the sentence text into an event-based BERT model to encode the input text $T$, the predicted event role $R$, and the embedding of \emph{Agent A} (used for dialogue generation described in Section \ref{sec:DialogueGeneration}) $EA=\{EA_{1}, EA_{2}, \dots, EA_{M}\}$. We extend BERT by adding a self-attention mechanism to learn new contextual representations of \emph{Agent A} (denoted as $SA$,) and input text (denoted as $ST$).

\subsubsection{\textbf{Final Representation}}
We concatenate the lexicon and event-based representation to produce the final representation $CT_{i}$, to be used for argument extraction. 
\begin{equation}
CT_{i}=LT_{i} \oplus ST_{i} \label{eq:r2}.
\end{equation}

\subsection{Ranked Argument Extraction}\label{sec:ArgumentRepresentation}

Given an event type, our argument extraction component aims to generate high-quality dialogue content by ranking the argument extraction order and utilizing the historical dialogue content.
It consists of four main modules: \emph{dialogue generation, argument extraction, incremental event learning, and reinforcement learning based dialogue management}.
The dialogue guided argument extraction model automatically extracts the arguments by inputting the actual event type and text. 
The incremental event learning module then adds the pseudo label as the training data and appends the pseudo-relationship to the lexicon-based graph.
After generating the target sentence's event representation, we follow a dialogue-guided strategy for argument extraction. Specifically, \emph{Agent A} uses the question set from \cite{DBLP:conf/emnlp/DuC20} to generate a query (or question) about an event or a chosen argument (e.g., ``What is the trigger of event \emph{X}?”). Then, \emph{Agent B} answers the query by predicting an argument role or event type. Based on the answer, \emph{Agent A} will then generate a new query in the next turn of argument extraction. This iterative process is driven by an RL-based dialogue management system described in Section \ref{Section 3.3}, aiming to optimize the order of argument extraction. The answer given by \emph{Agent B} will also be fed into an incremental event learning module described later to update the previous answers to be used for the next turn of argument extraction. Later in Table \ref{example}, we give an example dialogue generated by our approach. The automatically generated dialogue produces extra information to be used during the next-turn argument extraction. Our approach is highly flexible, allowing one to tailor the event extraction framework to a specific domain by populating the question set with domain knowledge.

\subsubsection{\textbf{Dialogue Generation}}\label{sec:DialogueGeneration}
Our dialogue generation module uses two agents (A and B) to assist event extraction through a sequence of question-answer conversations. Here, \emph{Agent A} generates dialogue content according to the currently processed role.
For each current role, it generates a question set \cite{DBLP:conf/emnlp/DuC20} to create more training data for argument extraction. 
For example, when the goal of \emph{Agent A} is to generate dialogue for the argument role``Instrument”, we select one of the pre-designed question set templates to generate dialogue. All we need to do is filling the argument roles for the given template.
\emph{Agent A} produces a hybrid content consisting of both the current argument role and arguments already extracted. 
\emph{Agent B} then generates the content, including the predicted argument given by the argument extraction (described in the next paragraph). 
The predicted argument is then fed into \emph{Agent A} to generate a new dialogue for the next turn of argument extraction. 
Similar to \emph{Agent A}, \emph{Agent B} also provides a dialog content generation template and only needs to fill the template with predicted arguments. If the predicted argument meets the confidence conditions, it will be part of the content of \emph{Agent A}.
The content of \emph{Agent B} will also be fed into the incremental event learning module described later to add high confidence results for the next turn of argument extraction.

\subsubsection{\textbf{Argument Extraction}}
\emph{Agent B} responds to a query by filling the answer slots of a simple answer template designed for each specific question template. It does so by using the learned representation, $CT_{i}$, to locate the start ($i_{s}$) and end ($i_{e}$) position of an argument within the target sentence. Specifically, we obtain the word probability of a chosen argument as:
\begin{equation}
{P}_{start}(r, t, k)=\frac{\exp \left(W^{r s} {CT}_{k}\right)}{\sum_{i=1}^{i=N} \exp \left(W^{r} {CT}_{i}\right)},\label{eq:r10}
\end{equation}
\begin{equation}
{P}_{end}(r, t, k)=\frac{\exp \left(W^{r e} CT_{k}\right)}{\sum_{i=1}^{i=N} \exp \left(W^{r} {CT}_{i}\right)},\label{eq:r11}
\end{equation}
where $W^{r s}$ and $W^{r e}$ are vectors to map ${CT}_{k}$ to a scalar. Each event type $t$ has a type-specific $W^{r s}$ and $W^{r e}$. 
The probability, ${P}_{span}(r, t, a_{i_{s}, i_{e}})$, of an argument being the description (or answer) for argument role $r$ and event type $t$ is given as follows:
\begin{equation}
{P}_{span}(r, t, a_{i_{s}, i_{e}})={P}_{start}(r, t, i_{s}) \times {P}_{end}(r, t, i_{e}). \label{eq:r12}
\end{equation}

\subsubsection{\textbf{Incremental Event Learning}}
Our incremental argument learning module tries to incorporate the information obtained at the current argument extraction turn to extract a new argument in the next turn. We do so by adding the extracted argument roles (i.e., pseudo labels) whose reward (evaluated by RL) is greater than a configurable threshold to the input text to provide additional information for extracting new arguments. We also add a new edge (i.e., pseudo relation) to connect the extracted arguments in the lexicon-based graph for an event, so that we can update the lexicon representation to be used for the dialogue in the next turn.

\subsubsection{\textbf{Reinforcement Learning-based Dialogue Management}}
\label{Section 3.3}

We use RL to optimize the argument extraction order during our iterative, dialogue-guided argument extraction process.

\textbf{Dialogue action.}
The dialogue-guided event extraction method defines the action as the set of event schema and argument. It indicates that the reinforcement learning algorithm requires determining the argument roles to go from the current argument to the following argument.
Different from the previous reinforcement learning-based method, we design two agents with varying spaces of action.
For \emph{Agent A}, action $\bm{a_{A}}$ is a role from the event schema, which is the action of \emph{Agent A}.
For \emph{Agent B}, action $\bm{a_{B}}$ is the argument, which is the action of \emph{Agent B}.
While it is essential to determine if the event type of current dialogue turn needs to be converted to the next event type. It means our method can determine well of event type changing.

\textbf{Dialogue state.}
Defined at time step $t$, state $S_{t} \epsilon S$, is characterized through $\bm{S_{t}=(R_{t}, c_{t}^{A}, c_{t}^{B}, Q_{0}, H^{C})}$.
Where $R_{t}$ is the argument role selected through the reinforcement learning-based dialogue management, $c_{t}^{A}, c_{t}^{B}$ are the current embedding of \emph{Agent A} and \emph{B}, $ Q_{0}$ is the initial question of \emph{Agent A} about event type, $H^{C}$ represents the history of the conversation.
Different historical conversation $H^C$ contributes differently to the target argument extraction.
States $s^{A}$ and $s^{B}$ are history-dependent, encoding the historical dialogue and $T$.
$s^{A}$ and $s^{B}$ are the concatenation of the state $s^{A}_{t-1}, s^{B}_{t-1}$ from the last dialogue content and the current content embedding $c^{A}_{t}, c^{B}_{t}$, represented as:
\begin{equation}
\mathrm{s}_{\mathrm{t}}^{A}=\mathrm{s}_{\mathrm{t}-1}^{B} \oplus \mathrm{c}^{A}_{\mathrm{t}},
\end{equation}
\begin{equation}
\mathrm{s}_{\mathrm{t}}^{B}=\mathrm{s}_{\mathrm{t}}^{A} \oplus \mathrm{c}^{B}_{\mathrm{t}}.
\end{equation}

\textbf{Dialogue policy network.}
The policy is choosing the right action for role selection.
The policy network is a parameterized probability map in action space and confidence degree, which aims to maximize the expected accumulated reward. 
\begin{equation}
\begin{array}{l}
\pi_{\theta}\left( a^{A}, a^{B} \mid s^{A}, s^{B})\right)= \pi\left(a^{A}, a^{B} \mid (s^{A}, s^{B}); \theta \right)\\
=\mathbb{P}\left(a^{A}_{t}=a^{A}, a^{B}_{t}=a^{B} \mid s^{A}_{t}=s^{A}, s^{B}_{t}=s^{B}, \theta_{t}=\theta\right),
\end{array} \label{eq:r13}
\end{equation}
where $\theta$ is the learnable parameter representing the weight on our dialogue policy network. 
The dialogue policy network decides that actions are chosen of $T$. It consists of two networks. The first network is the feed-forward network for encoding the dialogue histories is implemented using a softmax function. The second network is the BiLSTM for encoding the dialogue histories $\bm{H^{C}_{t}=(H^{C}_{t-1},a_{t-1}, s^{A}_{t}, s^{B}_{t})}$ being a continuous vector $h^{C}_{t}$, and $H^{C}_{t}$ is an observation, and $a_{t-1}$ is an action sequence, updated through BiLSTM.
\begin{equation}
e_{T}=\operatorname{softmax}\left(W_{T} F\left(s^{A}, s^{B}\right)+b_{T}\right), \label{eq:r14}
\end{equation}
\begin{equation}
h^{C}_{t} =BiLSTM\left(h^{C}_{t-1},\left[a_{t-1} ; s^{A}_{t}, s^{B}_{t}\right]\right), \label{eq:r15}
\end{equation}
where ${W}_{T}$ and ${b}_{T}$ are the parameters, $F\left({s}_{t}\right)$ is the state feature vector, $e_{T}$ is the vector of the input sentence $T$, $h^{C}_{t-1}$ is the representation in $t-1$ conversation, $a_{t-1}, s^{A}_{t}$ and $s^{B}_{t}$ are the action representation and the current state representation of \emph{Agent A} and \emph{B} respectively.

\textbf{Dialogue reward.}
The dialogue management module saves the historical dialogues. For the specific event type, the search space is limit. We design a reward function to evaluate all actions. The reward $\bm{R(s^{B}, a^{A}, a^{B})}$ is defined as the relatedness of the argument extraction part. 
\begin{equation}
R(s^{B}, a^{A}, a^{B})=\sum_{i}{P}_{span}(r_{i}, t, a_{i_{s}, i_{e}}) \label{eq:r15}.
\end{equation}
The policy agent can be effectively optimized using the reward signal. Note that we simultaneously identify all remaining arguments whose reward is less than a threshold in the final turn to avoid error propagation.
The threshold in our model is 0.75.

\subsection{Event Classification}\label{sec:EventClassification}
  \vspace{-1mm}
Event classification detects whether the input sentence is an event and classifies the event type to which the sentence belongs.
Each sentence is fed into a lexicon-based graph neural network model and an event-based BERT \cite{DBLP:conf/naacl/DevlinCLT19} model to learn the global knowledge and context knowledge of the sentence, respectively.
The event classification model detects what kinds of events the sentence contains by adding pseudo argument relation knowledge.
If the sentence does not contain an event, NULL is output, and the subsequent modules are not executed.
enabling it to distinguish the prediction error. 
We use a fully connected layer to compute the context-aware utterance representation $y_{i}$ as follows:
\begin{equation}
{y}_{i}=ReLU(\boldsymbol{W}({LT}_{i} \oplus {ST}_{i})+b), \label{eq:r3}
\end{equation}
where $\boldsymbol{W}$ and $b$ are trainable parameters, and $\oplus$ denotes vector concatenation operation.
For event classification sub-task, an $ReLU$ activation is used to enforce sparsity.
To improve event classification performance,
we design an extra task - predicting the number of event types.
Our multi-task event classification model calculates the combined loss of the two tasks to overcome the event type imbalance results in a low recall.

\subsubsection{\textbf{Trigger Classification}}
The existing event classification is based on the trigger identification to identify the event type, but our method is directly according to the input sentence to identify the event type. Therefore, when we evaluate the performance of trigger classification, we use the trigger predicted in the previous Section III-B of ranked argument extraction. We concatenate the predicted trigger and the input sentence to classify the event type.

\subsubsection{\textbf{Multi-task Joint Loss for Event Classification}}
The multi-task joint loss function estimates the difference in the predicted result and ground-truth value.
We design two tasks to learn the distinction between prediction error.
For the event classification task, we employ the cross-entropy loss function defined as:
\begin{equation}
\mathcal{L}_{T} =-\sum_{i} yt_{i} \log \left(\hat{yt}_{i}\right)+(1-yt_{i}) \log \left(1-\hat{yt}_{i}\right), \label{eq:r4}
\end{equation}
where $yt_{i}$ and $\hat{yt}_{i}$ are the $i$-th real label and the predicted label on the event classification task.
For the type number prediction task, we adopt the mean square loss with the $L_{2}$-norm:
\begin{equation}
\mathcal{L}_{N}=\sum_{i}\left\|\hat{yl}_{i}-yl_{i}\right\|^{2}+\eta\|\Theta\|_{2}, \label{eq:r5}
\end{equation}
where $yl_{i}$ and $\hat{yl}_{i}$ are the $i$-th label and prediction on the second task. $\Theta$ is model parameters, and $\eta$ is a regularization factor.
The overall loss function for optimizing the whole event classification model is:
\begin{equation}
L=\lambda_{1} L_{T}+\lambda_{2} L_{N}, \label{eq:r6}
\end{equation}
where $\lambda_{1}, \lambda_{2}$ are hyperparameters to balance the two loss. 
The $\lambda_{1}$ is 0.65 and $\lambda_{2}$ is 0.35 in our model.




\section{Experiments and Results}\label{sec:Experiments}

\begin{table*}[!htbp]
\centering
\caption{The P (precision), R (recall), and F1 score on event classification, argument identification and argument role classification results performed on the ACE 2005 test set. 
Best results are highlighted in bold and ``–" means results are not available. 
We use the two perspective event representation module and multi-turn dialogue module in our dialogue guided model.
The difference between our full model and dialogue-guided model is whether using reinforcement learning.
}
\label{ACE2005}
\resizebox{\textwidth}{!}{
\begin{tabular}{l|ccc|ccc|ccc|ccc|c}
\toprule
\textbf{Task} & \multicolumn{3}{c}{\textbf{Trigger Classification}} & \multicolumn{3}{c}{\textbf{Trigger Identification}} & \multicolumn{3}{c}{\textbf{Argument Identification}} & \multicolumn{3}{c}{\textbf{Argument Role Classification}} & \textbf{Runtime}  \\
 & P   & R   & F1 & P   & R  & F1   & P   & R   & F1 & P   & R  & F1  \\\midrule
DBRNN \cite{DBLP:conf/aaai/ShaQCS18} & 74.10  & 69.80  & 71.90 &- &- & - &71.30&64.50&67.70 & 66.20  & 52.80  & 58.70 &- \\
JMEE \cite{DBLP:conf/emnlp/LiuLH18} & 76.30  & 71.30  & 73.70 &80.20 &72.10 &75.90 &71.40&\textbf{65.60}&68.40& 66.80  & 54.90  & 60.30 &-\\
Joint3EE \cite{DBLP:conf/aaai/NguyenN19} & 68.00  & 71.80  & 69.80  &70.50 &74.50 &72.50 &59.90&59.80&69.90& 52.10  & 52.10  & 52.10  &-\\
GAIL-ELMo \cite{DBLP:journals/dint/ZhangJS19} &   74.80  & 69.40 & 72.0  &76.80 & 71.20& 73.90  &63.30&48.70& 55.10  & 61.60  &  45.70  &  52.40  &-  \\
PLMEE \cite{DBLP:conf/acl/YangFQKL19} &   81.00  & 80.40 & 80.70& 84.80 & 83.70 &84.20    &71.40&60.10& 65.30  & 62.30  &  54.20  & 58.00   &23.8h \\
Chen et al. \cite{DBLP:journals/corr/abs-1912-01586}& 66.70   &  74.70  & 70.50&68.90 &77.30 &72.90 &44.90 &41.20 &43.00 & 44.30 & 40.70  & 42.40 &20.3h \\
Du et al. \cite{DBLP:conf/emnlp/DuC20} &  71.12  &  73.70  &72.39 & 74.29& 77.42& 75.82 &58.9&52.08&55.29 & 56.77 & 50.24  &53.31   &24.6h  \\
MQAEE \cite{DBLP:conf/emnlp/LiPCWPLZ20} &  -  &  -  &71.70 &-&-& 74.50 &-&-&55.20 & - & - &53.40  &31.5h   \\
\hline 
\rowcolor{Gray} Our fine-tuned BERT &  75.34  & 76.15   & 75.93& 85.51 & 84.24 & 86.02 &70.42&60.24& 64.12 & 62.34 & 53.45  &  57.80  &32.9h   \\
\rowcolor{Gray} Our dialogue guided model &  78.24  & \textbf{80.44}  & 79.62 & 86.34 & 85.21 & 86.91 &72.75&63.53& 67.71 & 67.67 & \textbf{54.92}  &  59.42  &36.1h \\
\rowcolor{Gray} Our full model &  \textbf{81.23}  & 80.00   & \textbf{80.71}& \textbf{87.94} & \textbf{87.22} & \textbf{87.73} &\textbf{73.43}&65.30& \textbf{69.97} & \textbf{69.82} & 54.43  &  \textbf{61.42}  &38.1h \\\bottomrule
\end{tabular}
}
\end{table*}

\subsection{Tasks}

We test our approach on the ACE 2005 \cite{DBLP:conf/lrec/DoddingtonMPRSW04} dataset, which is the most widely-used dataset in event extraction. 
It contains 599 documents, annotated with 8 coarse-grained event types, 33 event subtypes, and 36 argument roles. 
The part that we use for evaluation is fully annotated with 5,272 event triggers and 9,612 arguments. To make our results directly comparable, we keep the same data split as previous work \cite{DBLP:conf/emnlp/LiuLH18, DBLP:conf/acl/YangFQKL19, DBLP:conf/emnlp/DuC20, DBLP:conf/emnlp/LiPCWPLZ20}. The number of documents for the training set, validation set, and test set is 529, 30, and 40, respectively.
It contains a complete set of training data in English, Arabic, and Chinese for the ACE 2005 technology evaluation. 
For the dialogue content generation of \emph{Agent A}, we generate content for the argument role selected by the reinforcement learning-based dialogue management module. For example, if the event type is "Life:Die" and the argument role is "instrument" for example in Fig. \ref{dialogue}, the generated content is ``What is the killing instrument of event Life:Die?". In dialogue, we add predicted arguments from \emph{Agent B}. To keep the sentence grammatically correct, we add fixed compositions, such as adding ``using" before the role of ``instrument". For the dialogue content generation of \emph{Agent B}, we generate content according to the predicted argument. Here, as with \emph{Agent A}, the template is pre-designed.
We evaluate the performance of our model and comparison models for trigger classification (TC), trigger identification (TI), argument identification (AI), and argument role classification (ARC) sub-tasks.
The evaluation metrics include precision (P), recall (R), and F1.

\subsection{Parameters}
We implement our model based on BERT 
\cite{DBLP:conf/naacl/DevlinCLT19}. We use BERT \cite{DBLP:conf/naacl/DevlinCLT19} as sequence encoding for queries and the hyperparameters of the decoder are the same as for the encoder. It has 12 layers, 768-dimensional hidden embeddings, 12 attention heads, and 110 million parameters.
The dialogue generation module generates multiple questions for the same trigger and argument. The final number of sentences is added to 2,4000 in training, validation, and test sets of 19,200, 2,400, and 2,400 sentences, at an 8:1:1 ratio.
The maximum sequence length is 512-word pieces, the learning rate is $3 \times 10^{5}$ with an Adam optimizer, the maximum gradient norm for gradient clipping is 1.0. The model is trained for 10 epochs and the batch size is 8, where we set the max question length to 128 and the max answer length to 64.
The optimal hyperparameters are tuned on the validation set by grid search, and we tried each hyperparameter five times.
The dialogue policy network contains a 128 unit bidirectional LSTM and a softmax layer. The dimension of BERT-based word embeddings is 512 dimensions. The mini-batch size is 128 in training. We use Adam as optimization algorithm with the gradient clipping being 5.

\subsection{Comparisons.}
We compare our extraction method with eight event extraction methods:
\textbf{DBRNN} \cite{DBLP:conf/aaai/ShaQCS18} leverages the dependency graph information to extract event triggers and argument roles.
\textbf{JMEE} \cite{DBLP:conf/emnlp/LiuLH18}, a jointly event extraction framework, introduces attention-based GCN to model graph information. 
\textbf{Joint3EE} \cite{DBLP:conf/aaai/NguyenN19} is a multi-task model that performs entity recognition, trigger detection and argument role assignment by shared Bi-GRU hidden representations. 
\textbf{GAIL-ELMO} \cite{DBLP:journals/dint/ZhangJS19} is an ELMo-based model that utilizes generative adversarial network to focus on harder-to-detect events.
\textbf{PLMEE} \cite{DBLP:conf/acl/YangFQKL19} is a BERT-based pipeline event extraction method and employs event classification depending on trigger.
 
\textbf{Chen et al.} \cite{DBLP:journals/corr/abs-1912-01586} use bleached statements giving models acquire to information included in annotation manuals.
\textbf{Du et al.} \cite{DBLP:conf/emnlp/DuC20} apply machine reading comprehension method employs event extraction and enhance data by constructing multiple question for each argument.
\textbf{MQAEE} \cite{DBLP:conf/emnlp/LiPCWPLZ20} is a multi-turn question answering method expediently utilizing history answer to implement event extraction.

\begin{figure*}[t]
    \centering
    \subfigure[GAT for different tasks on F1.]{
 \includegraphics[width=5.5cm]{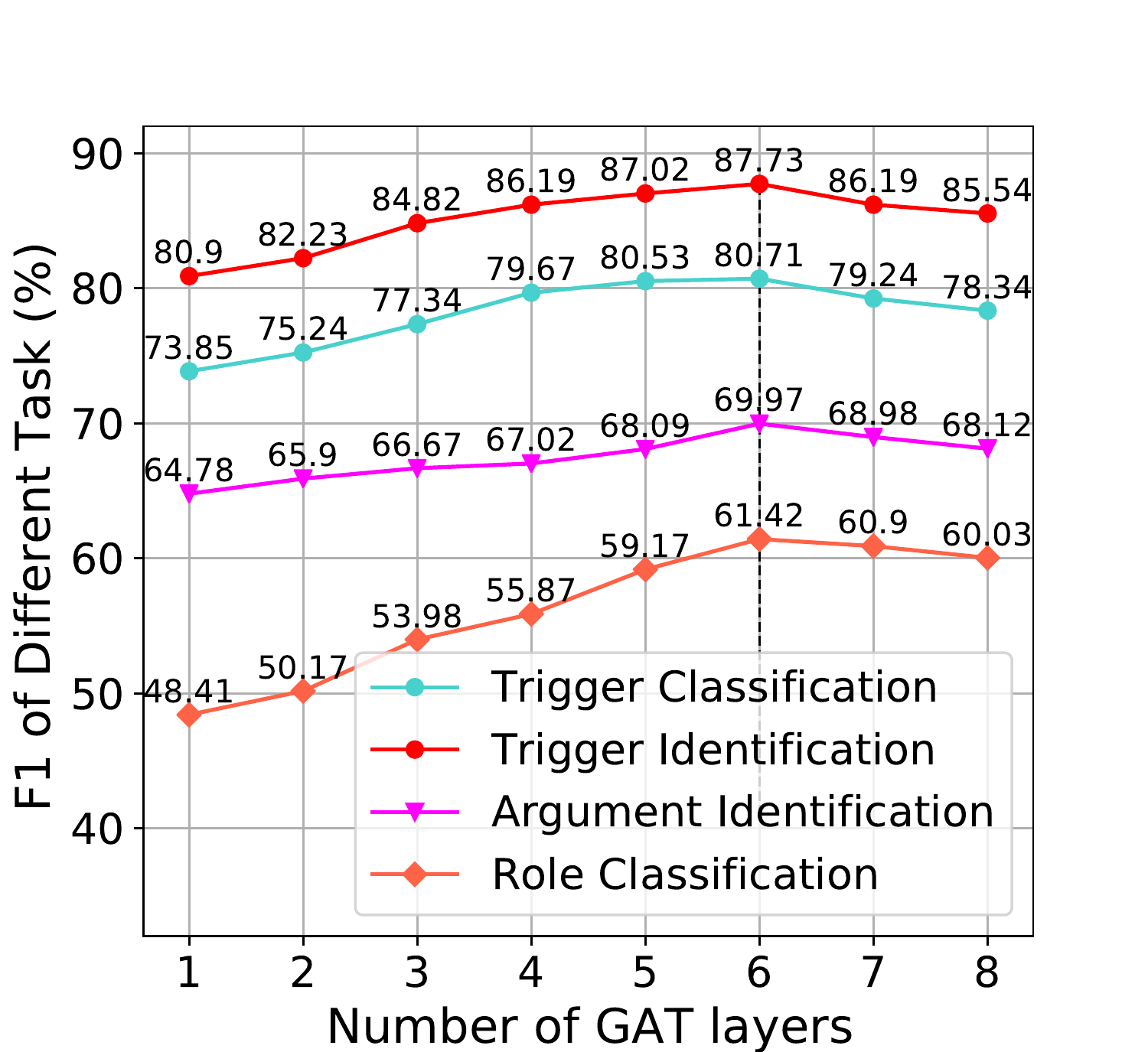}}
  \centering
 \subfigure[GAT for different tasks on precision.]{
 \includegraphics[width=5.5cm]{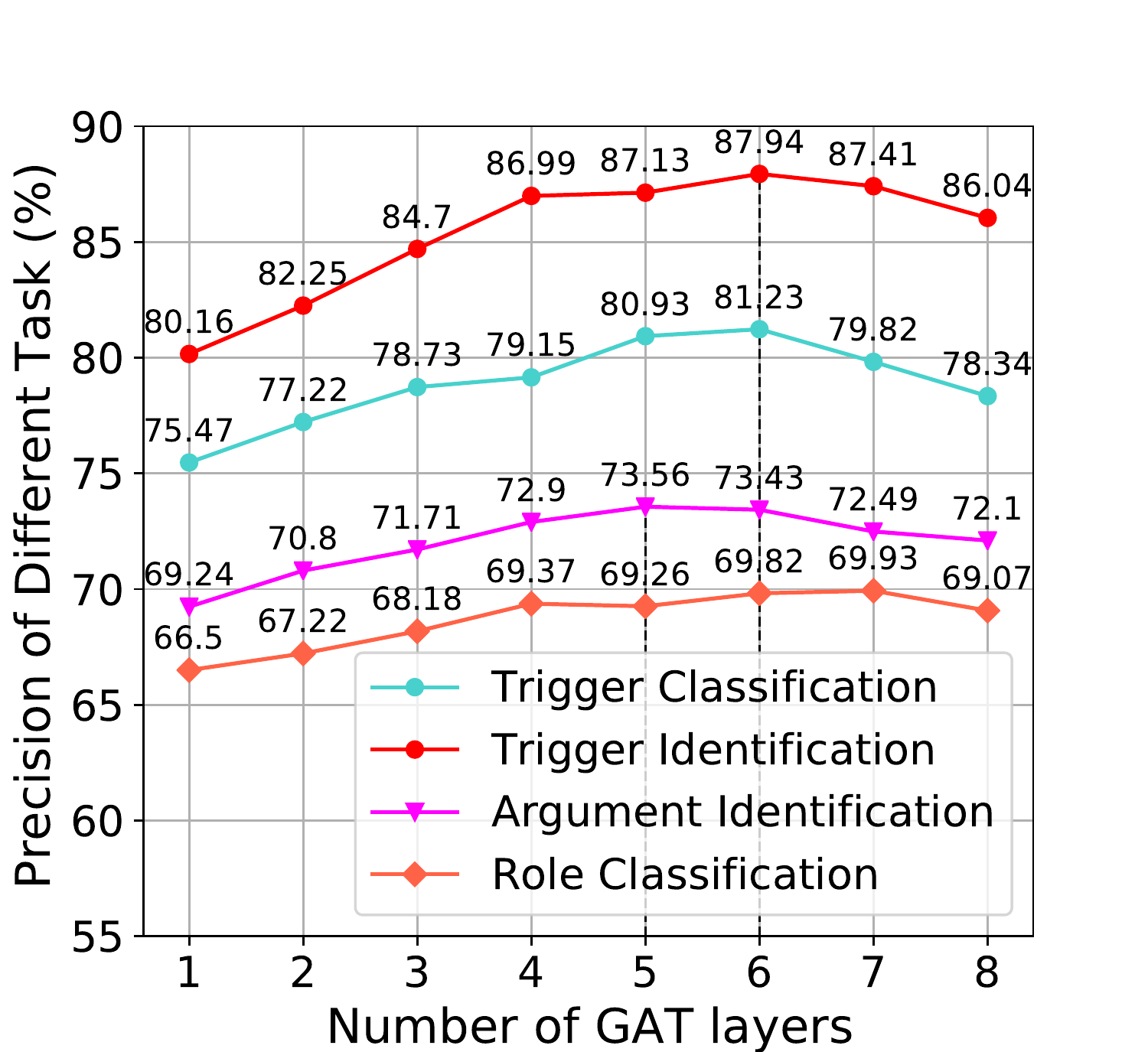}}
 \centering
 \subfigure[GAT for different tasks on recall.]{
 \includegraphics[width=5.5cm]{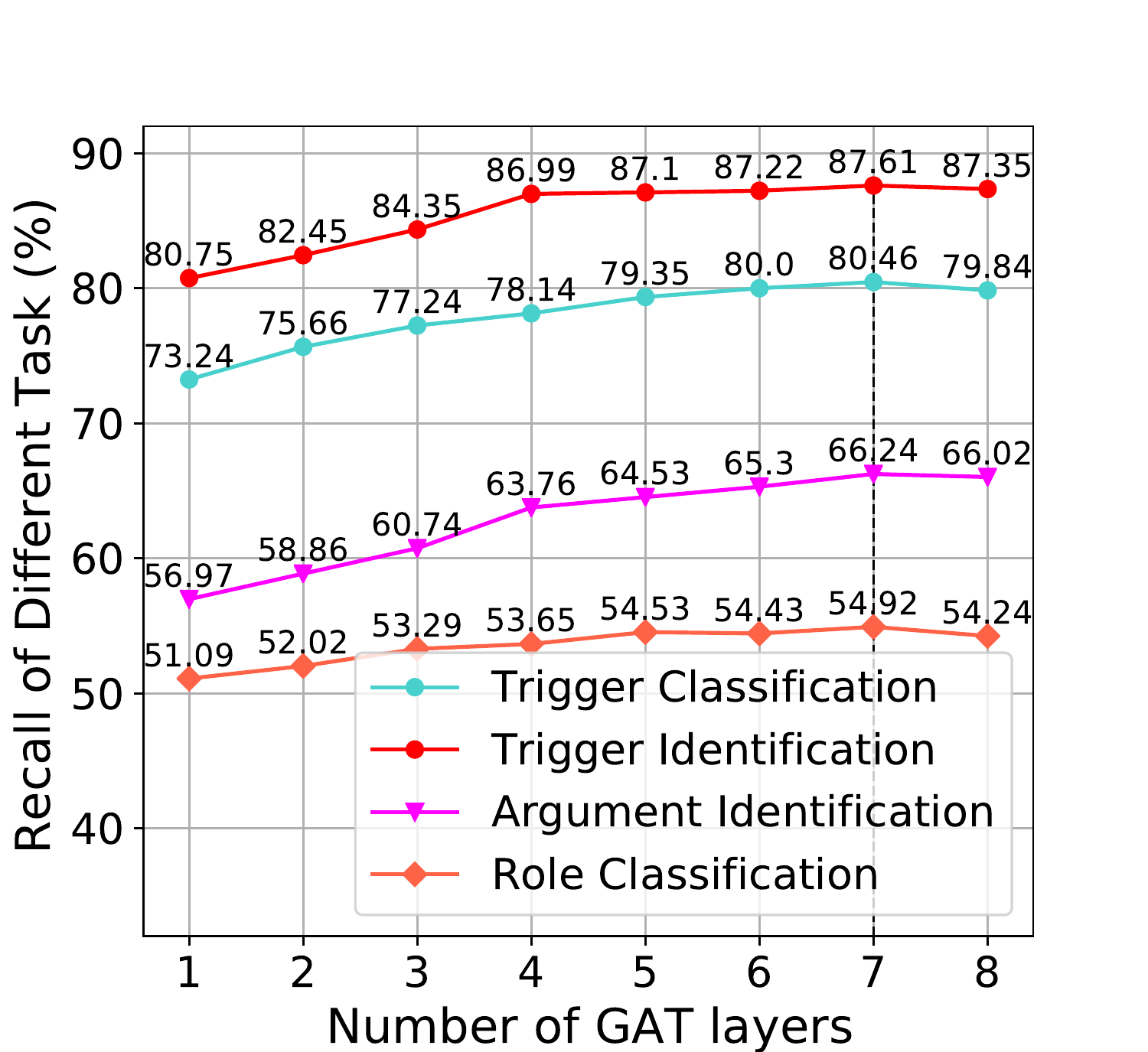}}
    \caption{The change of Precision, Recall and F1 for our model with different GAT layers under different tasks.}
    \label{GATtasks}
    \vspace{-2mm}
\end{figure*}

\begin{figure*}[t]
    \centering
    \subfigure[GAT for different label rate on TC.]{
 \includegraphics[width=5.5cm]{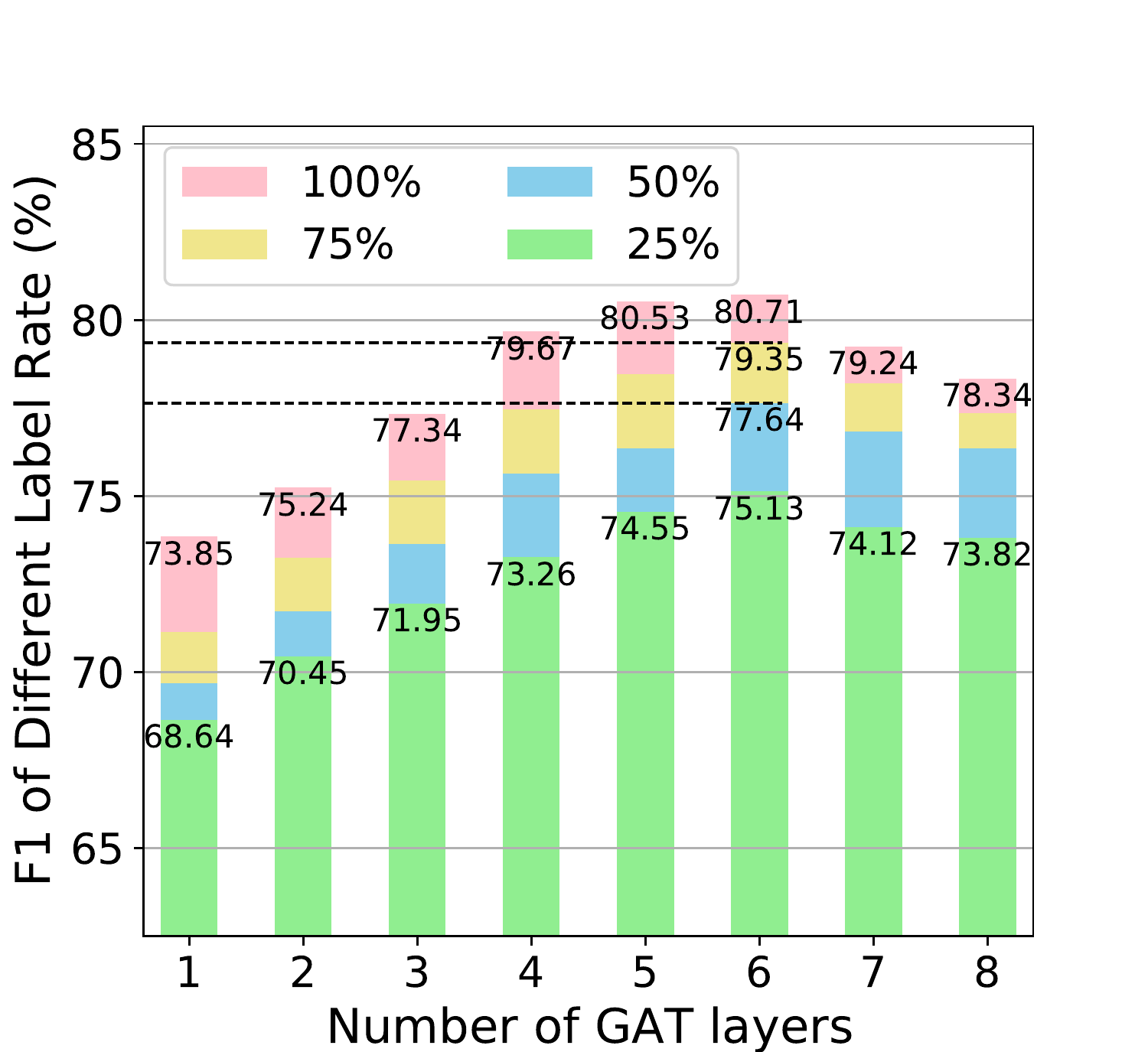}}
  \centering
 \subfigure[GAT for different label rate on AI.]{
 \includegraphics[width=5.5cm]{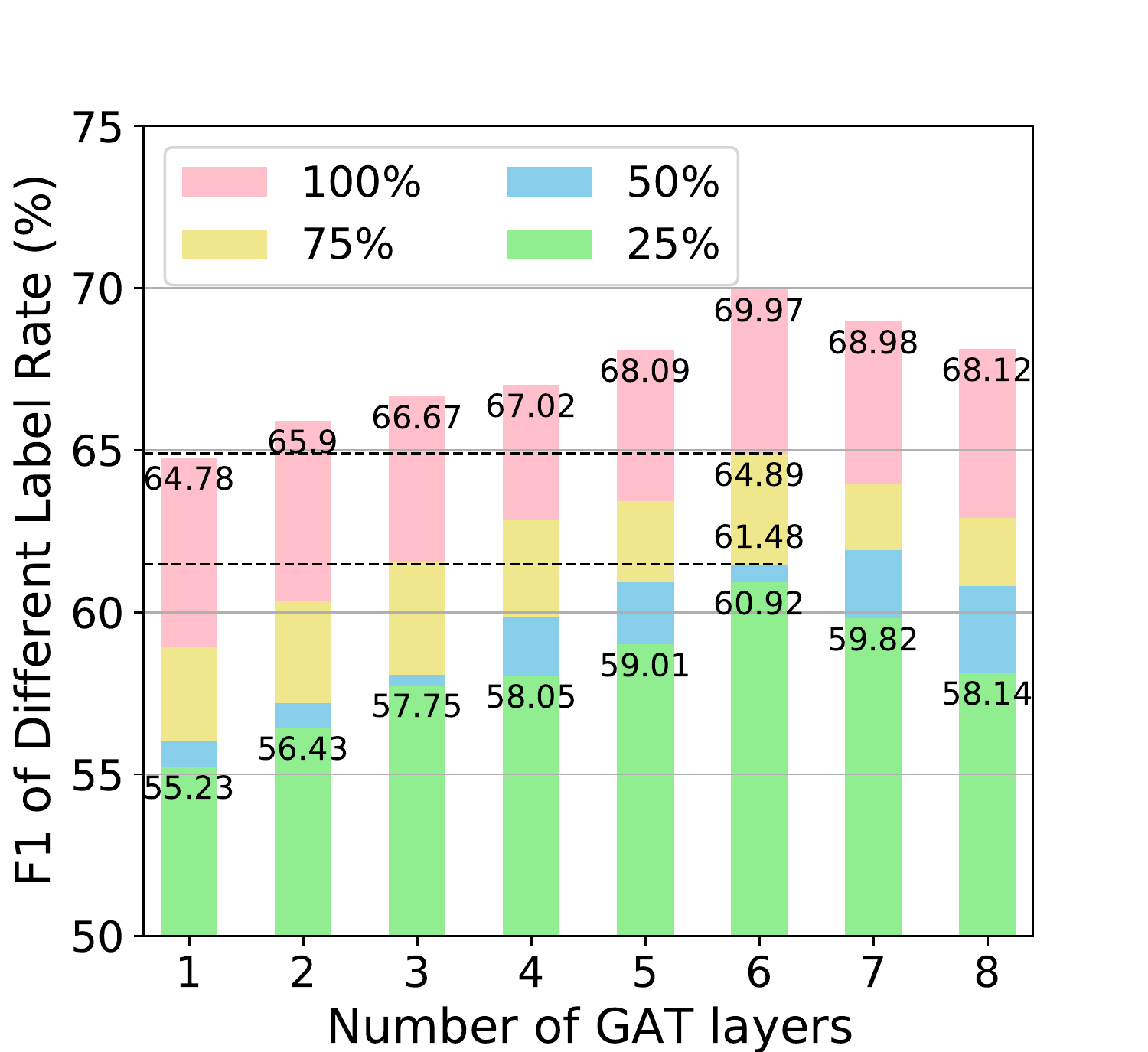}}
 \centering
 \subfigure[GAT for different label rate on RC.]{
 \includegraphics[width=5.5cm]{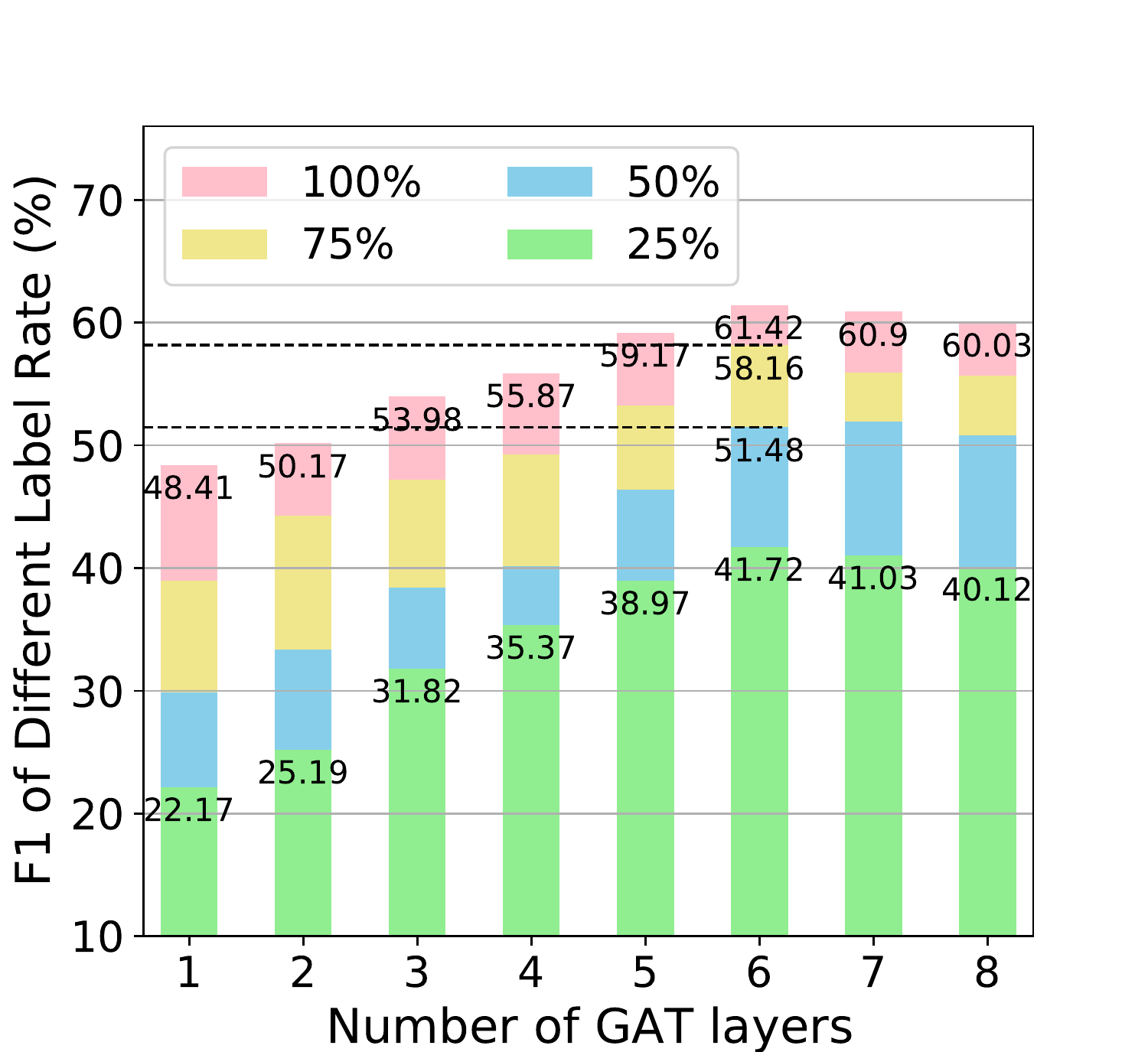}}
    \caption{The change of sub-tasks for our model with different GAT layers under different label rates.}
    \label{GATrate}
    \vspace{-2mm}
\end{figure*}

\subsection{Main Results}
Table \ref{ACE2005} shows the overall results of each approach per evaluation task with six GAT layers. 
In the trigger classification task, in order to make our model and baseline model adopt the same evaluation index, we supplement the trigger classification experiment. We use the predicted triggers together with the text as the input of the event classification model to predict the event type corresponding to the current triggers.
In our dialogue guided model, we use our event representation module and multi-turn dialogue model without reinforcement learning.
In our full model, we introduce reinforcement learning-based dialogue management.
The difference between full model and dialogue-guided model is whether using reinforcement learning to learn the order of argument extraction.
Our model consistently outperforms all other approaches on F1 and precision.
Compare to Du et al. \cite{DBLP:conf/emnlp/DuC20} and MQAEE \cite{DBLP:conf/emnlp/LiPCWPLZ20}, two recent machine reading comprehension models, our dialogue-guided model respectively achieves $7.23\%$ and $7.92\%$ improvements on the F1-score on the TC sub-task.
For TI and AI, our dialogue-guided model improves the F1-score by at least $10\%$ through reinforcement learning to guide the argument extraction order.
It achieves $6.11\%$ and $6.02\%$ improvements on the ARC sub-task.
It shows that our method is significantly superior to the MRC methods, which only utilize relationships among arguments.

\begin{figure*}[t]
    \centering
    \subfigure[Impact of labelled data ratio.]{
 \includegraphics[width=5.5cm]{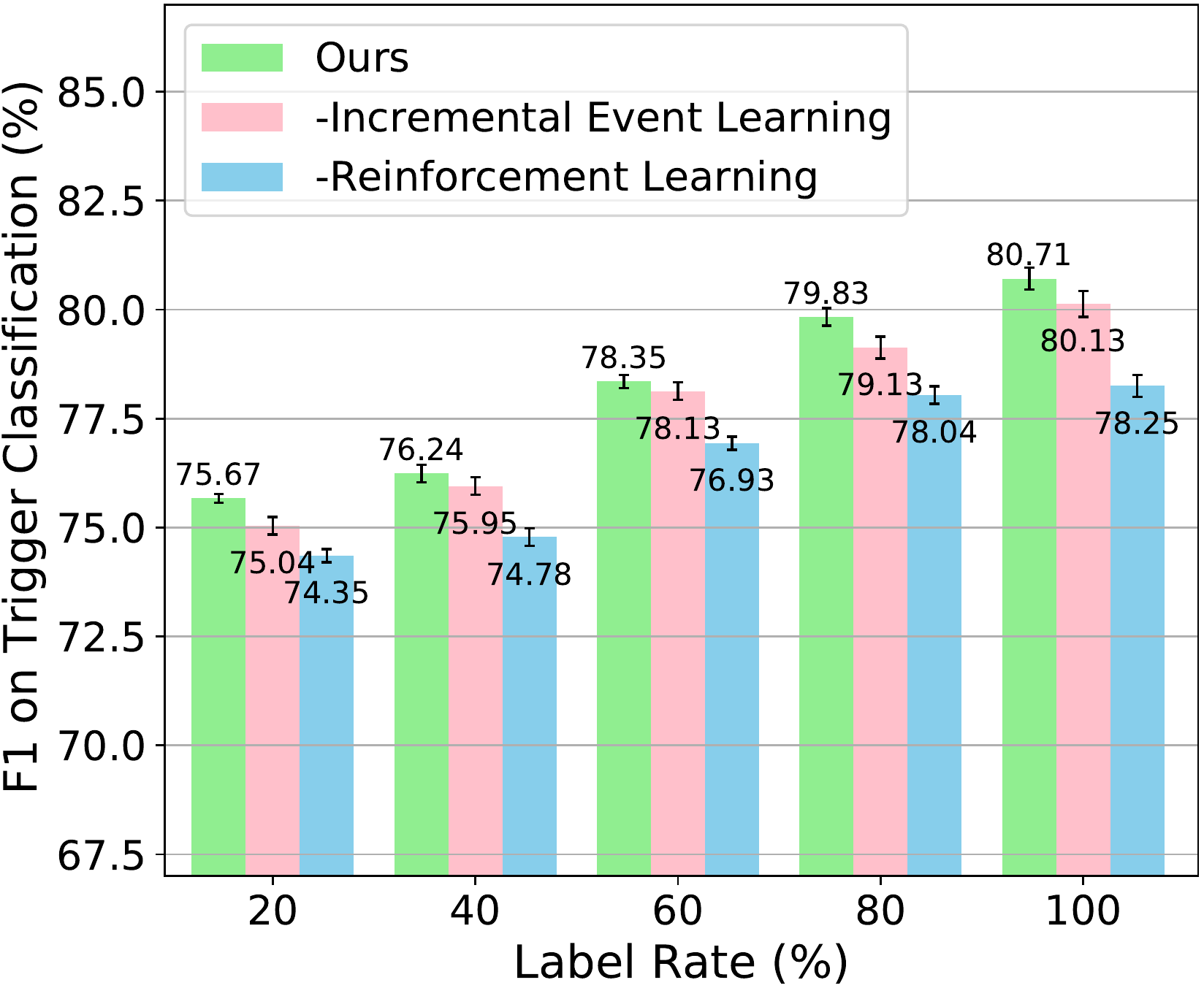}}
  \centering
 \subfigure[Impact of labelled data ratio.]{
 \includegraphics[width=5.5cm]{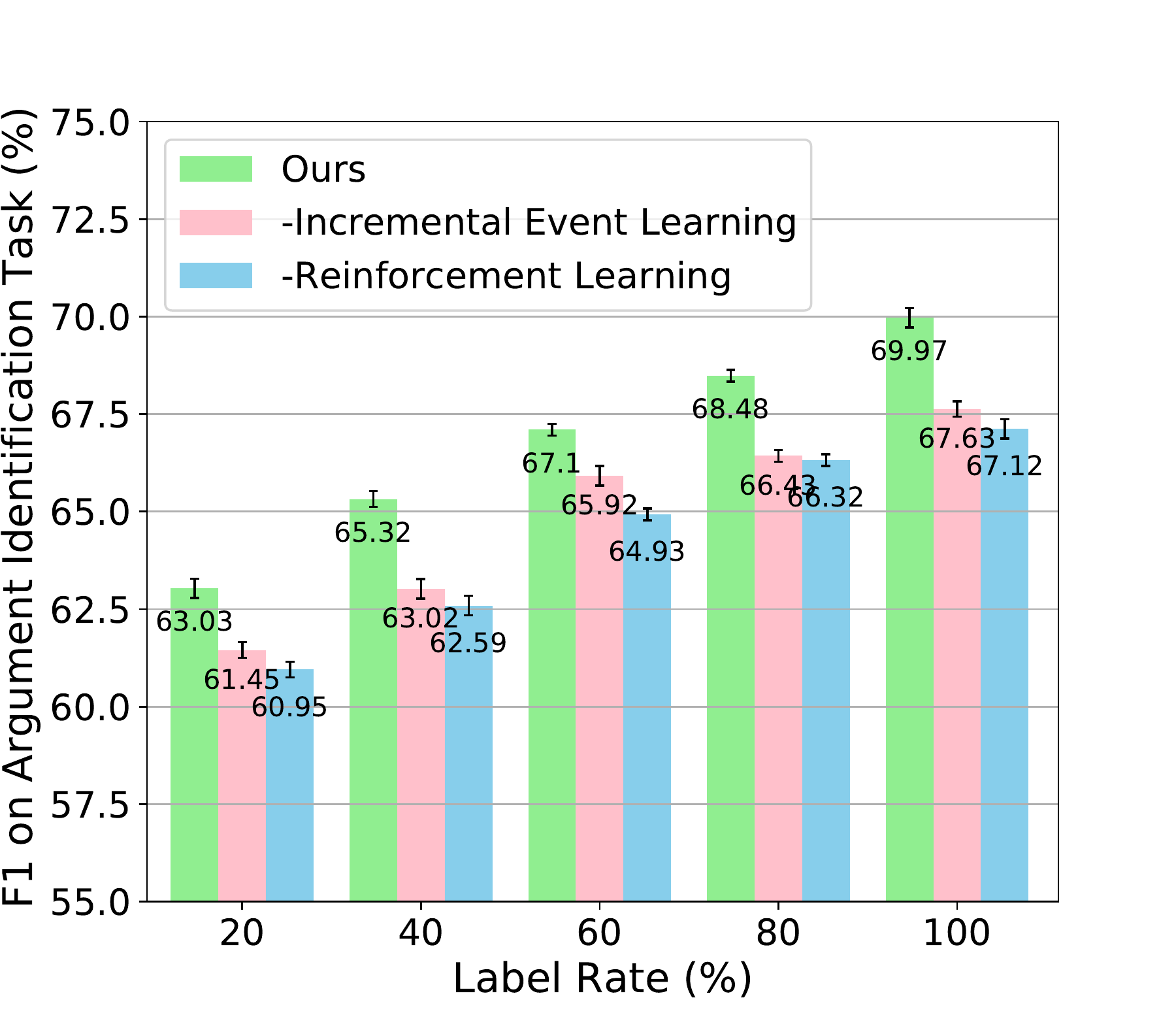}}
   \centering
 \subfigure[Impact of labelled data ratio.]{
 \includegraphics[width=5.5cm]{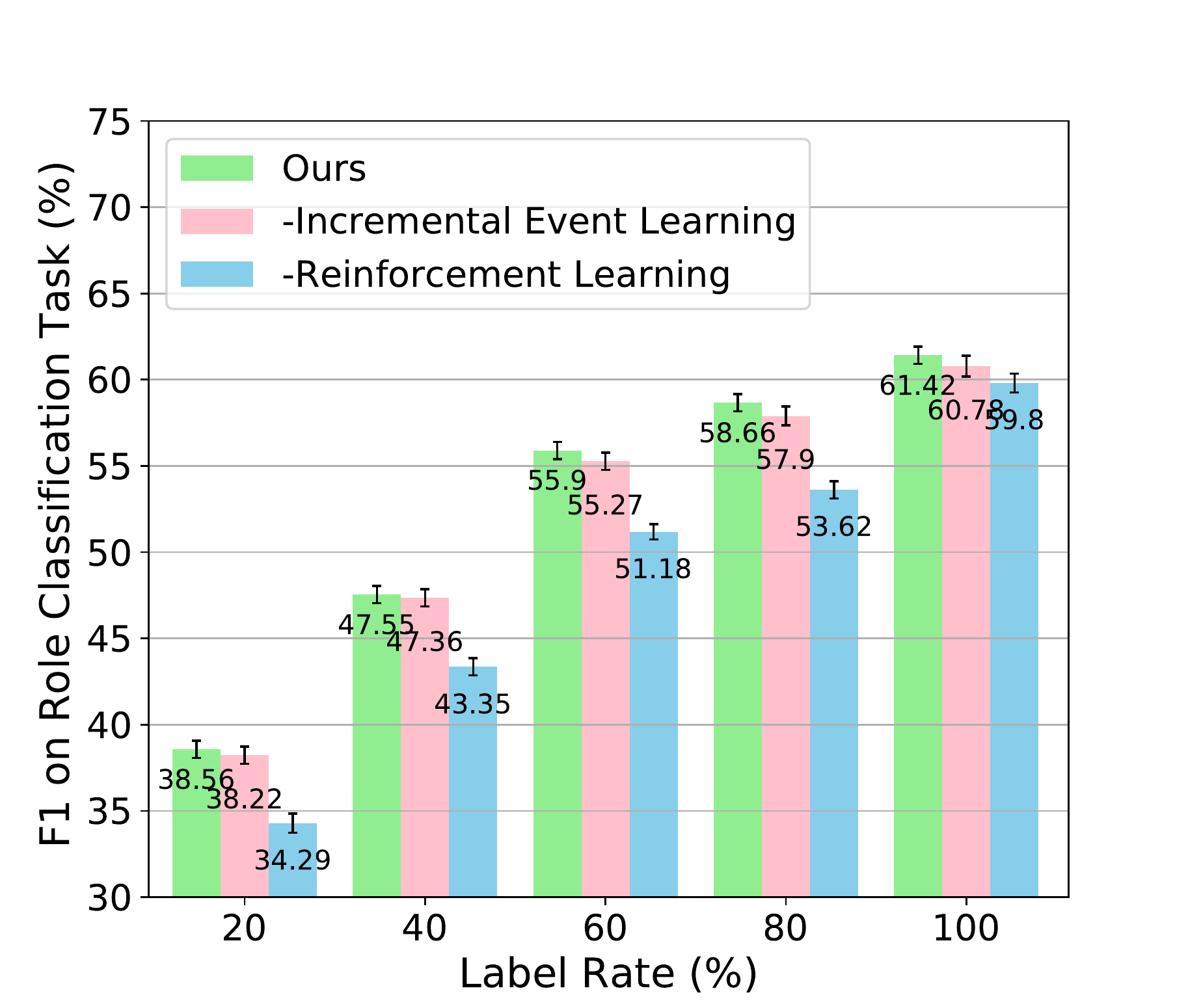}}
    \caption{The resulting F1 score of our approach as the ratio of labeled data changes for different tasks on different tasks.}
    \label{rate2}
\end{figure*}

\begin{figure*}[t]
    \centering
    \subfigure[Impact of argument roles.]{
 \includegraphics[width=5.5cm]{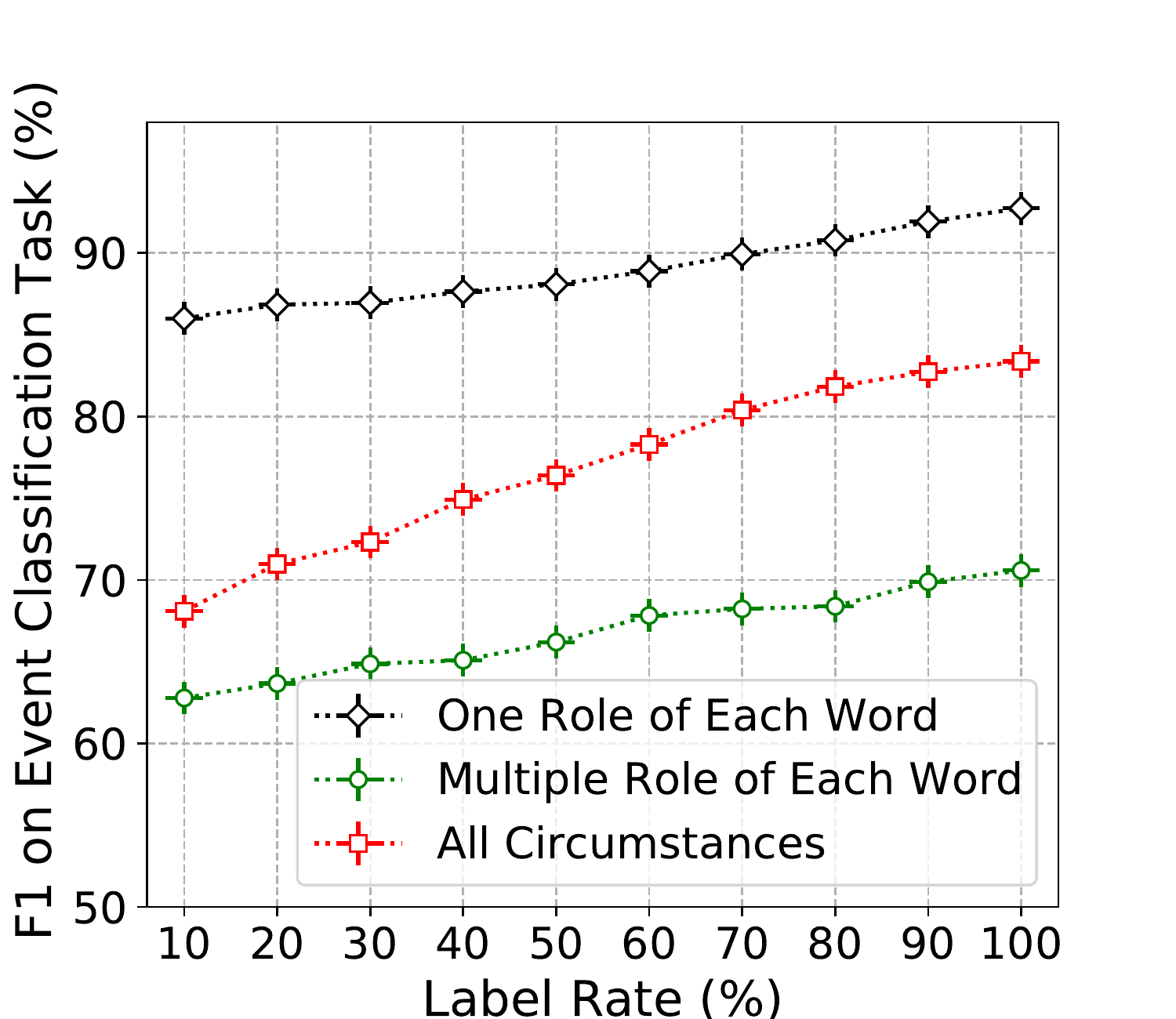}}
  \centering
 \subfigure[Impact of argument roles.]{
 \includegraphics[width=5.5cm]{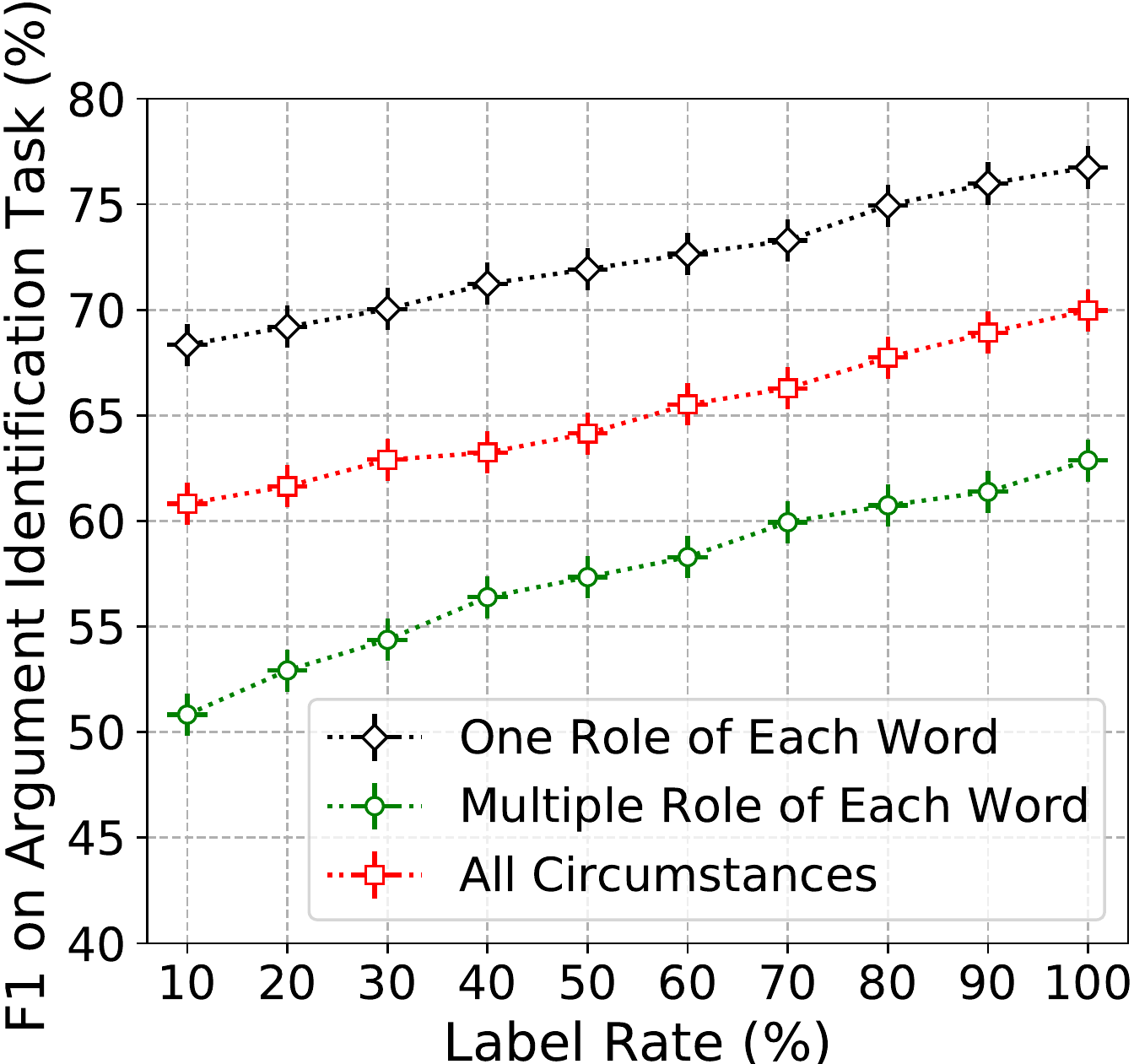}}
   \centering
 \subfigure[Impact of argument roles.]{
 \includegraphics[width=5.5cm]{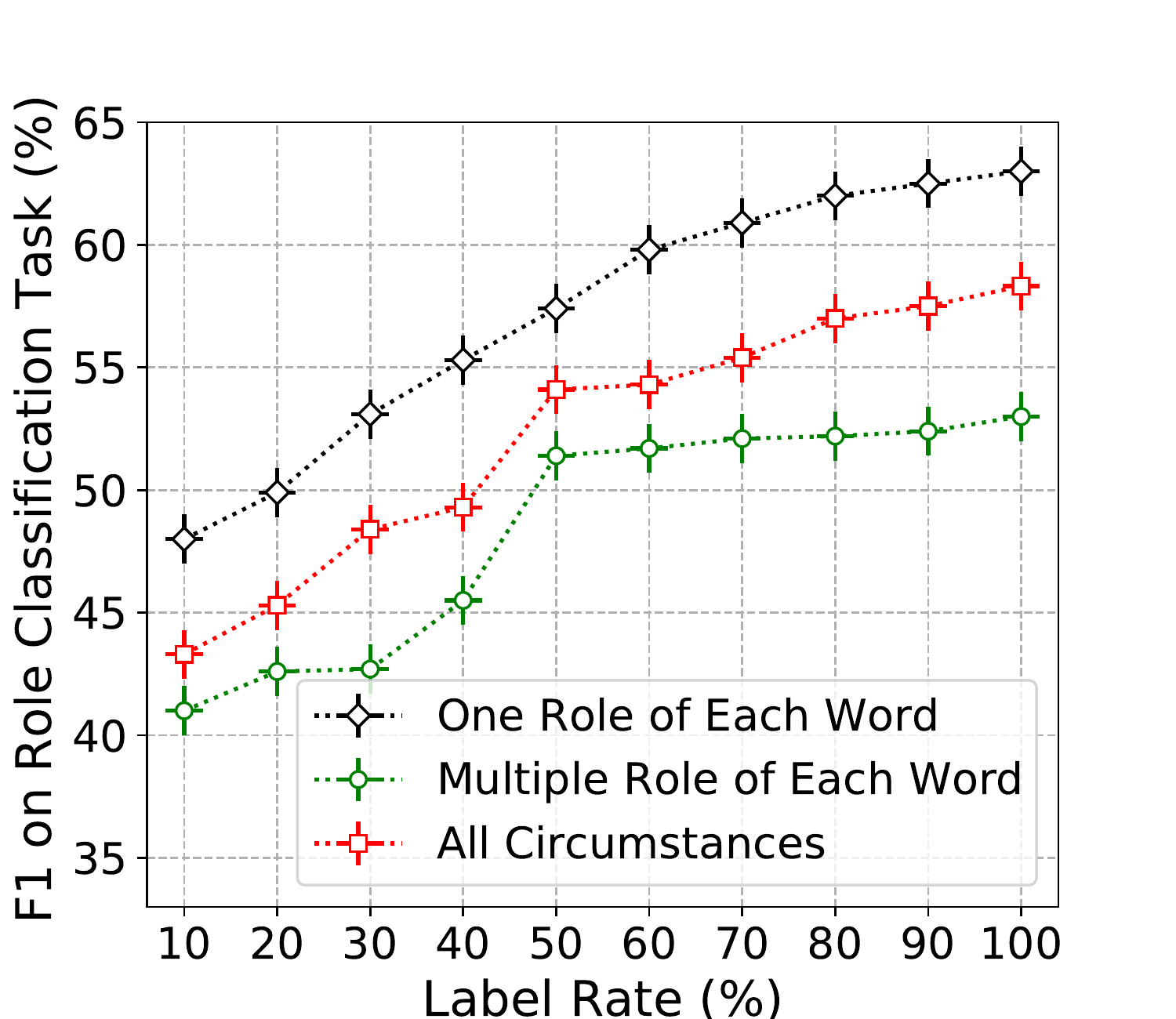}}
    \caption{The F1-score of our approach as the ratio of labeled data changes for multiple type test data on different tasks.}
    \label{rate_split}
       \vspace{-2mm}
\end{figure*}

\begin{figure*}[t]
    \centering
    \subfigure[Average loss on different iteration.]{
 \includegraphics[width=5.5cm]{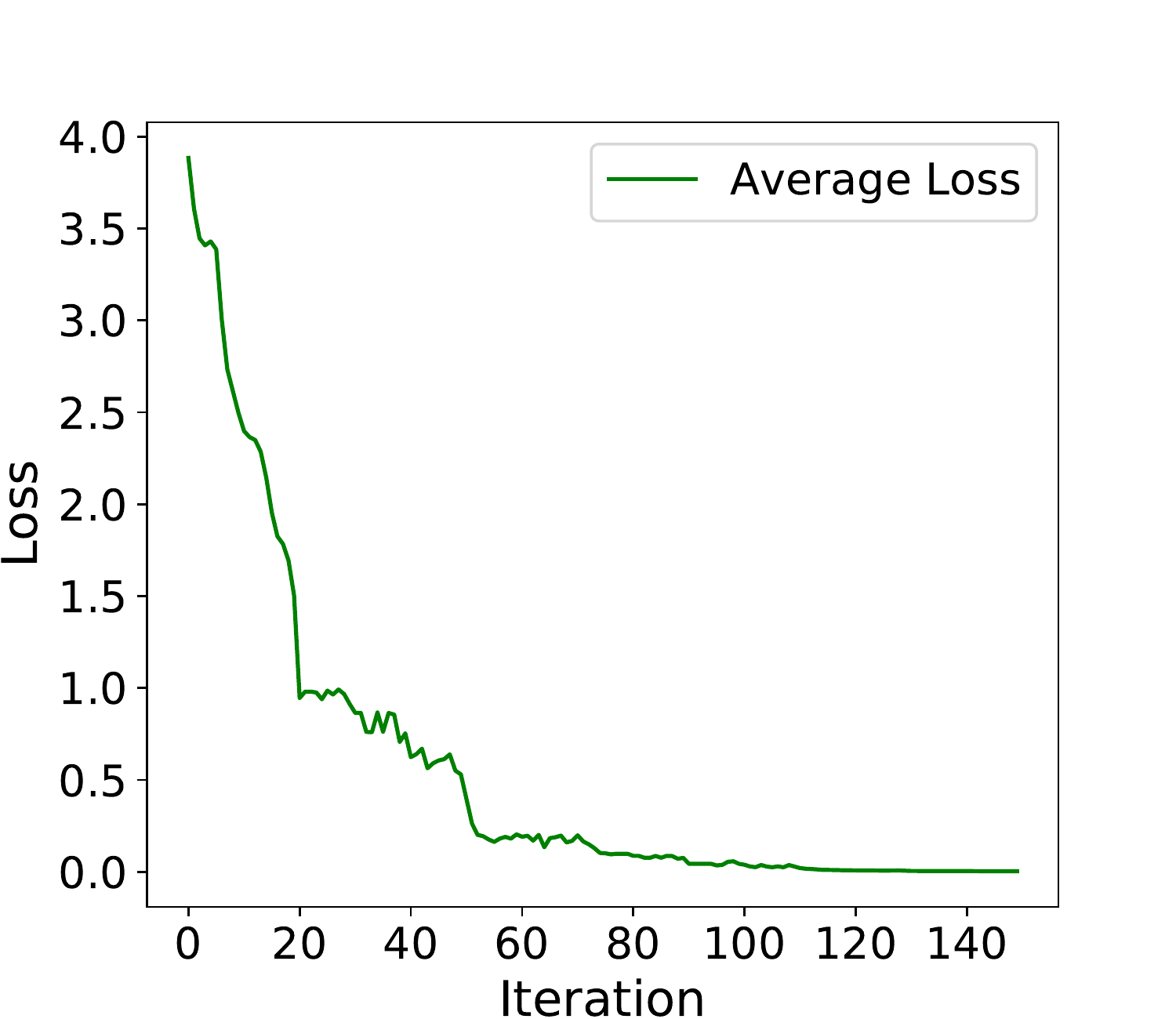}}
  \centering
 \subfigure[Average reward on different iteration.]{
 \includegraphics[width=5.5cm]{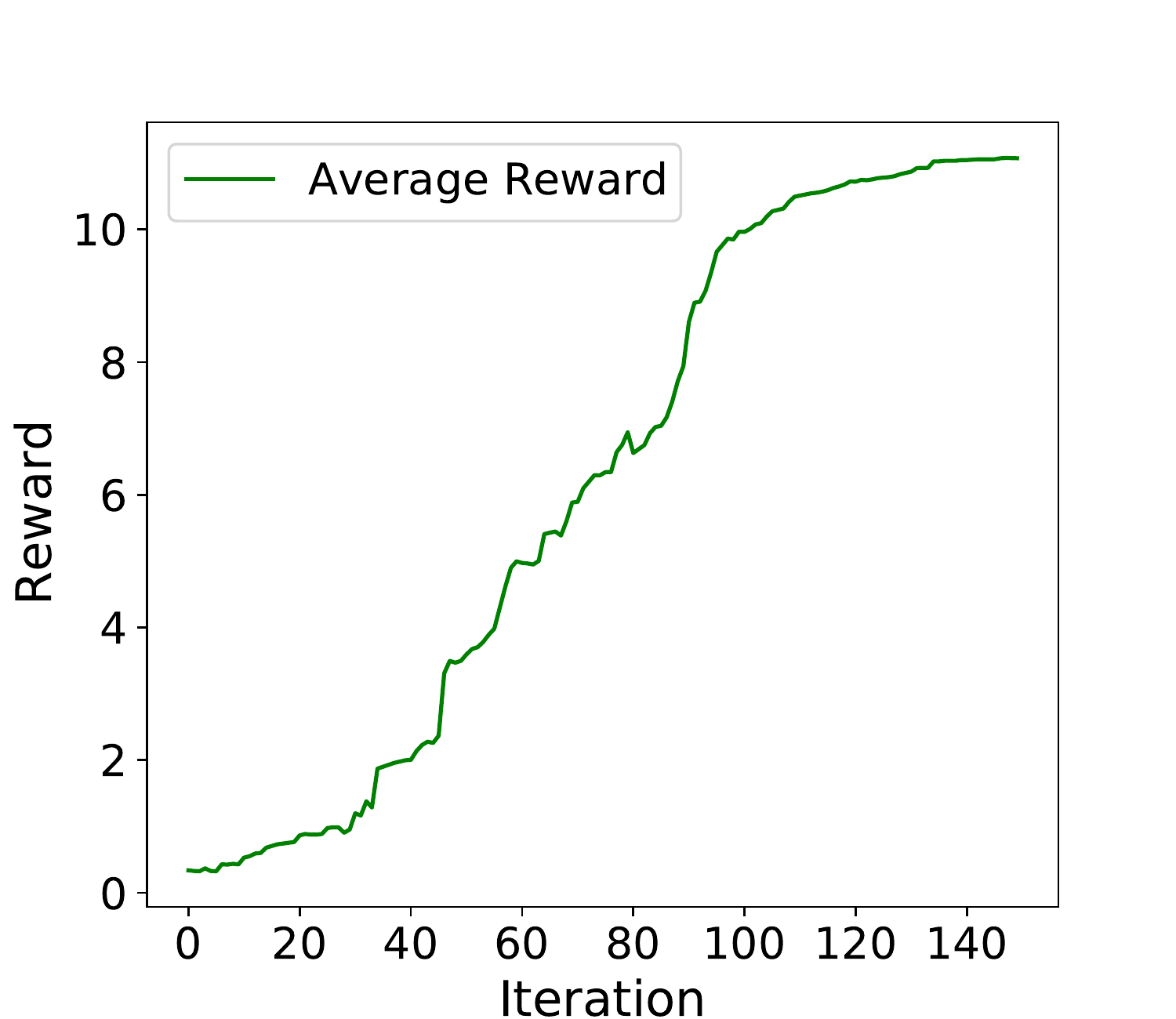}}
   \centering
 \subfigure[F1 on different iteration.]{
 \includegraphics[width=5.5cm]{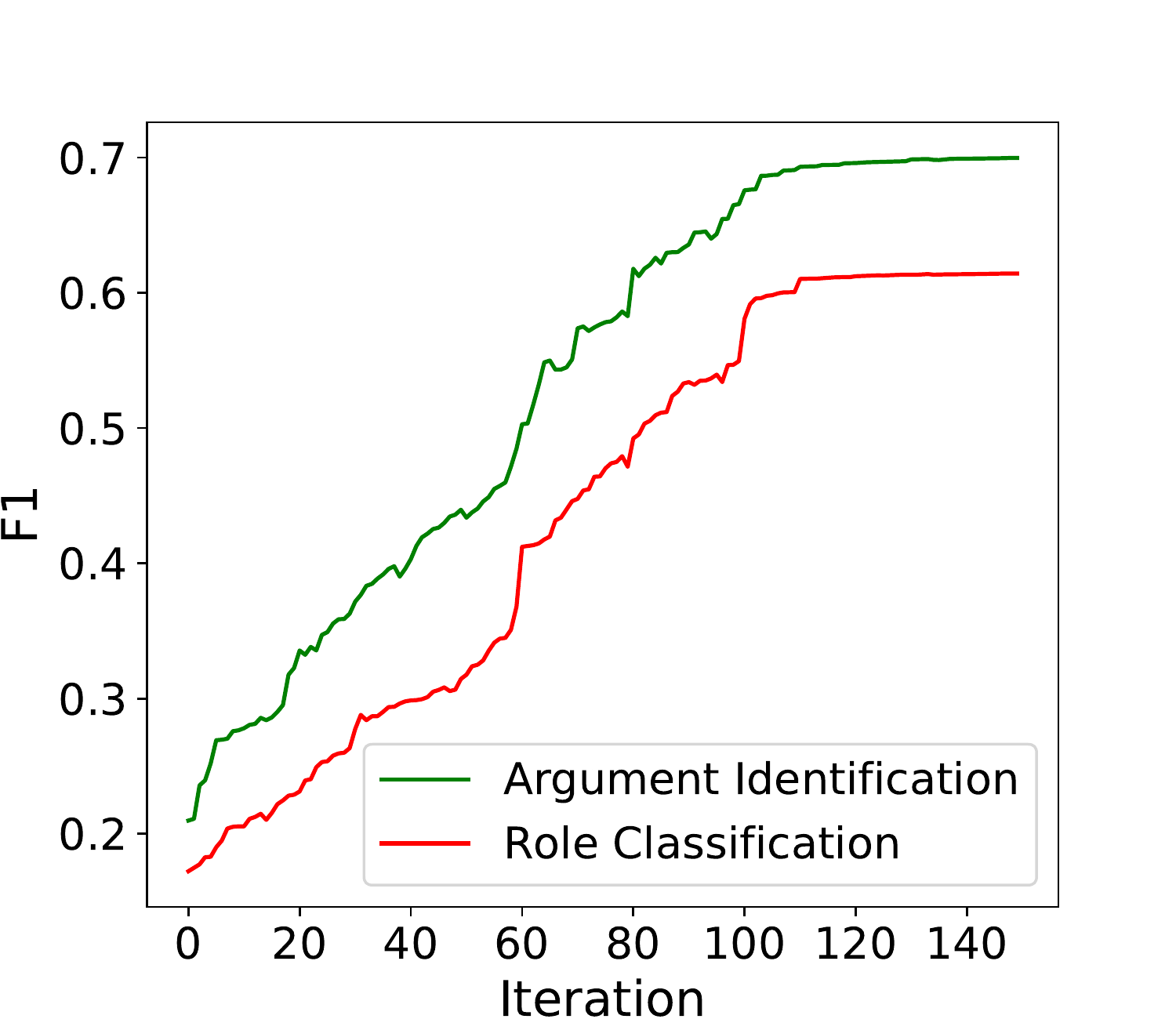}}
    \caption{The influence of reinforcement learning on argument extraction task.}
    \label{RL}
       \vspace{-2mm}
\end{figure*}

Our dialogue guided model also consistently outperforms PLMEE \cite{DBLP:conf/acl/YangFQKL19}, the best-performing baseline model. The results show the importance of exploiting argument knowledge and relation for event classification and argument extraction. Our full model boosts the F1-score by $9.01\%$, $13.23\%$, $14.77\%$, and $8.02\%$ on TC, TI, AI, and RC, respectively, when compared to the MQAEE, the best-performing alternative on F1-score. Our approach delivers higher precision than other approaches. On some tasks, our approach gives a lower Recall compared to the best-performing baseline model, but the resulting Recall is not far from the best model JMEE ($0.47\%$) and our dialogue guided model exceeds it. By utilizing argument extraction order, our approach delivers the best overall results.

We compare the runtime of our model with BERT-based baseline models, including the QA-based model. In our model, event extraction is realized through multiple rounds of our dialogue system, and reinforcement learning is introduced to optimize the argument extraction order. It can accomplish event extraction more accurately but increases training complexity.
The original intention of our model is to improve the precision, recall and F1 of event extraction by making full use of the dependence among arguments under limited data. We will consider improving the accuracy without increasing the model complexity and training difficulty in future work.

\subsection{Impact of GAT Layer Number}

A GAT model in a semi-supervised task with a lower label rate, which means the proportion of labeled data to the ACE dataset, requires more graph attention layers to maintain the best performance.
Fig~\ref{GATtasks} presents some empirical evidence to demonstrate the layer effect on our lexicon-based graph for all sub-tasks. 
We test the performance on TC, TI, AI, and RC four tasks and observe that the F1-score enhances when the GAT layer increases until the sixth layer reach the maximum. With respect to Precision and Recall, GAT achieves the best performance on the sixth and seventh layer.

As shown in Fig~\ref{GATrate}, we also test our model's performance with different layers in special label rates for TC, AI and RC sub-tasks.
It is apparent to note that the model gets the best performance under different label rates when the GAT layer is 6, and the number of layers under the best performance exhibits an increasing trend as the label rate increases.
It demonstrates that the best GAT layer gets a stable performance under different tasks and changing the label rate.
As can be seen from Fig.~\ref{GATtasks} and Fig.~\ref{GATrate}, using more GAT layers can improve the performance, judging by the Precision, Recall and F1-score. However, the improvement reaches a peak when using 6 GAT layers across all three evaluated tasks, and a further increase in the number of layers does not give improved performance. Therefore, we choose to use 6 GAT layers.

\subsection{Impact of Data Settings}
To verify that our model can achieve significant performance improvement even with little labeled data, we test the model's performance by changing the training data amount.
We evaluate the impact of training data on three subtasks in Fig~.\ref{rate2}.
Compared with the other two sub-tasks, trigger classification is the least affected by the labelled data ratio, which can basically reach 75\% F1-score. It shows that our model can achieve relatively stable results under different data scales.
Fig.~\ref{rate2}c shows how the F1-score changes as the ratio of labeled data available to our scheme changes on argument role classification task. As expected, using more labelled data thus improves the performance of our models. Our reinforcement learning module can achieve good performance, even when the amount of labeled data is small. The combination of reinforcement and incremental event learning can improve the robustness of our approach, leading to better overall results than individual techniques. 
Still, when we remove the incremental event learning and reinforcement learning module, respectively, the former changes dramatically than the latter.
It indicates that the incremental event learning module can make our model insensitive to the data scale change.

\begin{table*}[t]
\centering
\caption{Ablation study on global constraints on F1-score (\%).}
\label{Ablation}
\resizebox{\textwidth}{!}{
\begin{tabular}{l|rr|rr|rr|rr}
\toprule

\textbf{Tasks} & \multicolumn{2}{|c}{\textbf{Trigger Classification} }& \multicolumn{2}{|c}{\textbf{Trigger Identification}} & \multicolumn{2}{|c}{\textbf{Argument Identification}} &  \multicolumn{2}{|c}{\textbf{Argument Role Classification}} \\ 
 &  P  & F1  & P  & F1 & P  & F1& P  & F1 \\
\midrule
 \textbf{Ours}  &\textbf{81.23}& \textbf{80.71}   &\textbf{87.94} &\textbf{87.73} & \textbf{73.43}& \textbf{69.97}&\textbf{69.82}& \textbf{61.42}   \\ 
 \textbf{ -RLD} &79.34&78.25 &86.21&86.33&71.82&67.12 &  67.92 &59.80     \\ 
\textbf{ -IEL}   &79.62& 80.13&87.01&86.85&71.92& 67.63 &67.99& 60.78       \\ 
\textbf{ -LGAT}  & 80.52& 80.46 &86.21&86.10&72.03& 69.45 &68.51&61.01\\  
 \textbf{ -MTL} &80.02 &79.49  &86.93&86.81&71.89&  67.72  &67.20&60.22 \\
\bottomrule
\end{tabular}
}
\end{table*}

\begin{table*}[!t]
\centering
\caption{A dialogue generated by our RL-based dialogue system for sentence: ``\emph{As the \textbf{soldiers} approached, the \textbf{man} \textbf{detonated} \textbf{explosives} in the car, \textbf{killing} all four of the soldiers.}".}
\label{example}
\footnotesize
\begin{tabularx}{\linewidth}{Xl}
\toprule
\textbf{\emph{Agent A} query} & \textbf{\emph{Agent B} answer}\\
\midrule

\rowcolor{Gray}\textbf{A:} What is the first \textbf{event type}? & \textbf{B:} The first event type is \textbf{Life: Die}.  \\    

\textbf{A:} What is the trigger of event \textbf{Life:Die}? & \textbf{B:} The \textbf{trigger} of event \textbf{Life:Die} is \textbf{killing.}     \\ 

\rowcolor{Gray} {\textbf{A:} What is the \textbf{killing instrument} of event \textbf{Life:Die}?} & \textbf{B:} The \textbf{killing instrument} of event \textbf{Life:Die} is \textbf{explosives}.     \\ 

{\textbf{A:} Who is the \textbf{killing victim} of event \textbf{Life:Die} using  \textbf{explosives}?}  & {\textbf{B:} The \textbf{killing victim} of event \textbf{Life:Die} is the \textbf{soldiers}.} \\

\rowcolor{Gray} {\textbf{A:} What \textbf{agent} killing the \textbf{soldiers} of  event \textbf{Life:Die} using \textbf{explosives}?} &  {\textbf{B:} The \textbf{agent} of event \textbf{Life:Die} is the \textbf{man}.}\\

{\textbf{A:} What is the second \textbf{event type}?} & \textbf{B:} The second event type is \textbf{Conflict: Attack.}  \\

\rowcolor{Gray} {\textbf{A:} What is the \textbf{trigger} of event \textbf{Conflict: Attack}? }& \textbf{B:} The \textbf{trigger} of event  \textbf{Conflict: Attack} is \textbf{detonated.}     \\

\makecell[l]{\textbf{A:} What is the \textbf{detonated instrument} of event \textbf{Conflict: Attack}?} &\makecell[l]{\textbf{B:} The \textbf{detonated instrument} of event \textbf{Conflict: Attack} is \textbf{explosives} and \textbf{car}.}   \\

\rowcolor{Gray}  {\textbf{A:} What is the \textbf{detonated target} of event \textbf{Conflict: Attack} using \textbf{explosives} and \textbf{car}?} &\textbf{B:} The \textbf{detonated target} of event \textbf{Conflict: Attack} is \textbf{soldiers}.     \\

 {\textbf{A:} Who is the \textbf{detonated attacker} of event \textbf{Conflict: Attack} using \textbf{explosives} and \textbf{car}?}  & {\textbf{B:} The \textbf{detonated attacker} of event \textbf{Conflict: Attack} is \textbf{man}.} \\
\bottomrule 
\end{tabularx}
\end{table*}

Fig.~\ref{rate_split} shows the impact of argument compositions. In this experiment, we divide the test data into three parts where each word has (1) one role (\emph{One Role of Each Word}), (2) more than one role (\emph{Multiple Role of Each Word}) and (3) a mixture of both (\emph{All Circumstances}). 
Our model achieves 92.74\%, 76.74\% and 63.00\% F1-score on the part of one role of each word.
As expected, our approach gives better results when the argument has just one role than other data compositions. Nonetheless, our framework still delivers good performance for other data settings and can use the increased labeled data to improve performance. 
For sentences having multiple roles of words, our model is less affected by the label rate in the event classification task, but more affected by the label rate in the argument extraction sub-task. In general, our model can have relatively stable performance in both one role and multiple roles for each word.

\begin{figure*}[!t]
    \centering
    \includegraphics[width=0.9\linewidth]{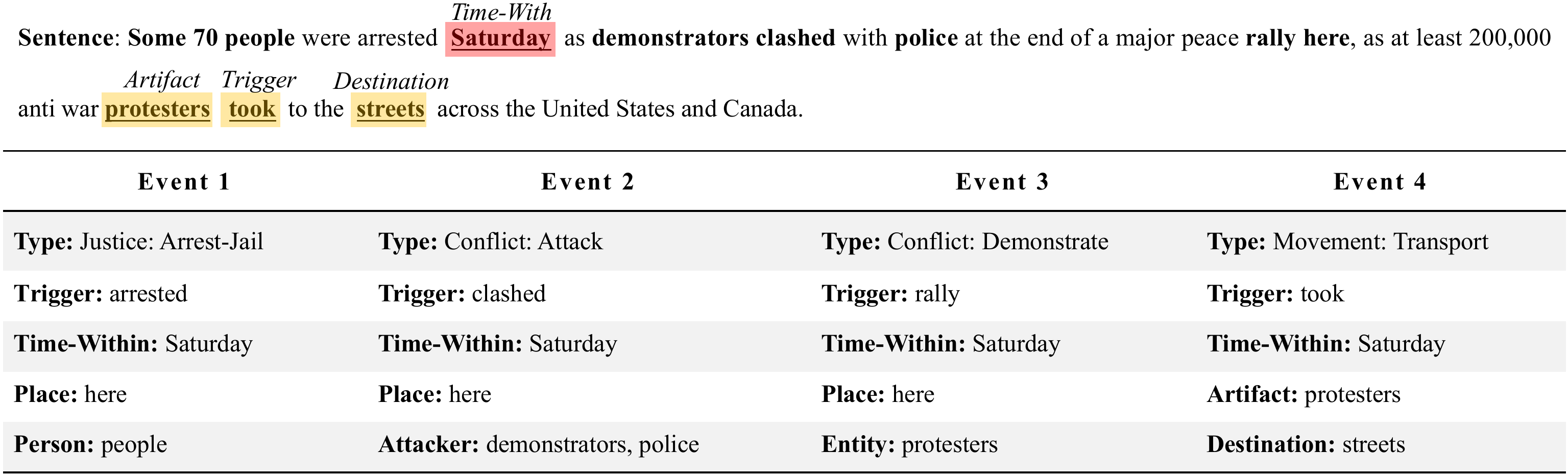}
    \caption{An example on ACE 2005 with a standard answer table for event extraction. The words in bold are the event arguments.}
    \label{caseStudy}
      \vspace{-5mm}
\end{figure*}

\subsection{The influence on RLD module}
In order to verify the reinforcement learning helping for argument extraction, and the model performance is improved through iterative training. We demonstrate this from three aspects: average loss, average reward, and F1 score. Our model converges in the $110th$ iteration, with a loss value of 0.0282. The average reward of reinforcement learning converges in the $120th$ iteration and is 10.7256. The F1 score or argument identification and role classification converge in the $120th$ iteration. The F1 argument identification subtask is 0.6142, and the F1 of role classification is 0.6997. Therefore, our model converges on $120th$ iterations when we add the RLD module.

\subsection{Ablation Study}

We evaluate four variants of our approach, given in Table \ref{Ablation}. We remove the \textbf{R}einforcement \textbf{L}earning-based \textbf{D}ialogue (\textbf{RLD}) module, which is the most key module.
The descending on TI, AI and RC sub-tasks is $1.40\%$, $2.85\%$ and $1.62\%$ F1-score, leading to performance changing significantly.
When it comes to the \textbf{I}ncremental \textbf{E}vent \textbf{L}earning (\textbf{IEL}) module,
F1-score decreases $0.58\%$, $0.88\%$, $2.34\%$ and $0.64\%$, respectively.
It may prove that our IEL module can provide useful pseudo labels and relations to model the arguments relation.
We remove the \textbf{L}exicon-based \textbf{G}raph \textbf{At}tention network (\textbf{LGAT}) module.
F1-score decreases on all four sub-tasks, which shows that the LGAT module positively affects learning lexicon-based knowledge. It may prove that our lexicon-based graph attention network and event-based BERT model can learn better of the event representation.
Moreover, the \textbf{M}ulti-\textbf{T}ask \textbf{L}earning (\textbf{MTL}) module is employed for TC. It improves $1.22\%$ F1 score by distinguishing the error types and contributes F1 score of TI in terms of $0.92\%$, AI in terms of $2.25\%$ and ARC in terms of $1.20\%$.
It suggests that the MTL module can accurately identify the sentence with multiple same event types. 
The results suggest that all variants are useful, and the RLD is the most important, as removing it can result in the most drastic performance degradation. 
By utilizing historical dialogue knowledge, our approach achieves about $6.55\%$, $11.83\%$, $11.92\%$ and $6.40\%$ F1-score gains than the best-reported question answering based method MQAEE \cite{DBLP:conf/emnlp/LiPCWPLZ20} on the four subtasks.
It may demonstrate the effectiveness of reinforcement learning-based dialogue generation.
For argument role classification, the precision enhances $7.52\%$ compared with the best-reported model PLMEE \cite{DBLP:conf/acl/YangFQKL19}.
It may prove that our lexicon-based graph attention network and event-based BERT model can learn better of the event representation.

\subsection{Case Study}

Table \ref{example} gives a dialogue conversation generated by our approach (Section \ref{sec:ArgumentRepresentation}). 
Our model can solve arguments with multiple roles, including words with different roles in different contexts. It can recognize the argument more completely by learning the relationship between arguments and the order of argument extraction. For example, in the sentence ``As the soldiers approached [...]", the word ``soldiers" can play different roles. When the event type is ``Life: Die", its role is ``Victim", while when the event type is
"Conflict: Attack", its role becomes ``Target". By contrast, the word ``explosives" plays the same role (``Instrument") in different events. Both cases can be correctly recognized by our approach.

In some scenarios where the text contains multiple events, our approach may fail to identify some arguments of scattered distribution. For example, as shown in Fig~\ref{caseStudy}, the sentence ``Some 70 people were arrested Saturday [...]" contains four event types: ``Justice: Arrest-Jail", ``Conflict: Attack", ``Conflict: Demonstrate" and ``Movement: Transport". For the latest event type, our model does not recognize ``Saturday" although it has the same role in multiple events. This is because this argument is further away from other arguments, and our model does not explicitly capture such relation when determining the event extraction order. Our future work will look into this.

\vspace{-2mm}
\section{Conclusion}\label{sec:Conclusion}
  \vspace{-1mm}
We have presented a new approach for event extraction by utilizing event arguments' relationships. We tackle the problem within a task-oriented dialogue guided framework designed for event extraction. Our framework is driven by reinforcement learning. We use RL to decide the order for extracting arguments of a sentence, aiming to maximize the likelihood of successfully inferring the argument role. We then leverage the already extracted arguments to help resolve arguments whose roles would be difficult to settle by considering the argument in isolation. Our multi-turn event extraction process also uses the newly obtained argument information to update decisions of the previously extracted arguments. This dual-way feedback process enables us to exploit the relation among event arguments to classify the argument's role in different text contexts. We evaluated our approach on the ACE 2005 dataset and compared to 7 prior event extraction methods. Experimental results show that our approach can enhance event extraction, outperforming competing methods in the majorities of the tasks. In the future, we plan to improve the multi-semantic representation of the dialogue guided event extraction by introducing commonsense knowledge.

\section*{Acknowledgment}
We thank the anonymous reviewers for their insightful comments and suggestions. 
Jianxin Li is the corresponding author.
The authors of this paper were supported by the NSFC through grant (No.U20B2053), the ARC DECRA Project (No. DE200100964), and in part by NSF under grants III-1763325, III-1909323, III-2106758, and SaTC-1930941.

\ifCLASSOPTIONcaptionsoff
  \newpage
\fi

\footnotesize
\bibliographystyle{ieeetr}
\bibliography{TASLP}

\begin{IEEEbiography}[{\includegraphics[width=1in,height=1.25in,clip,keepaspectratio]{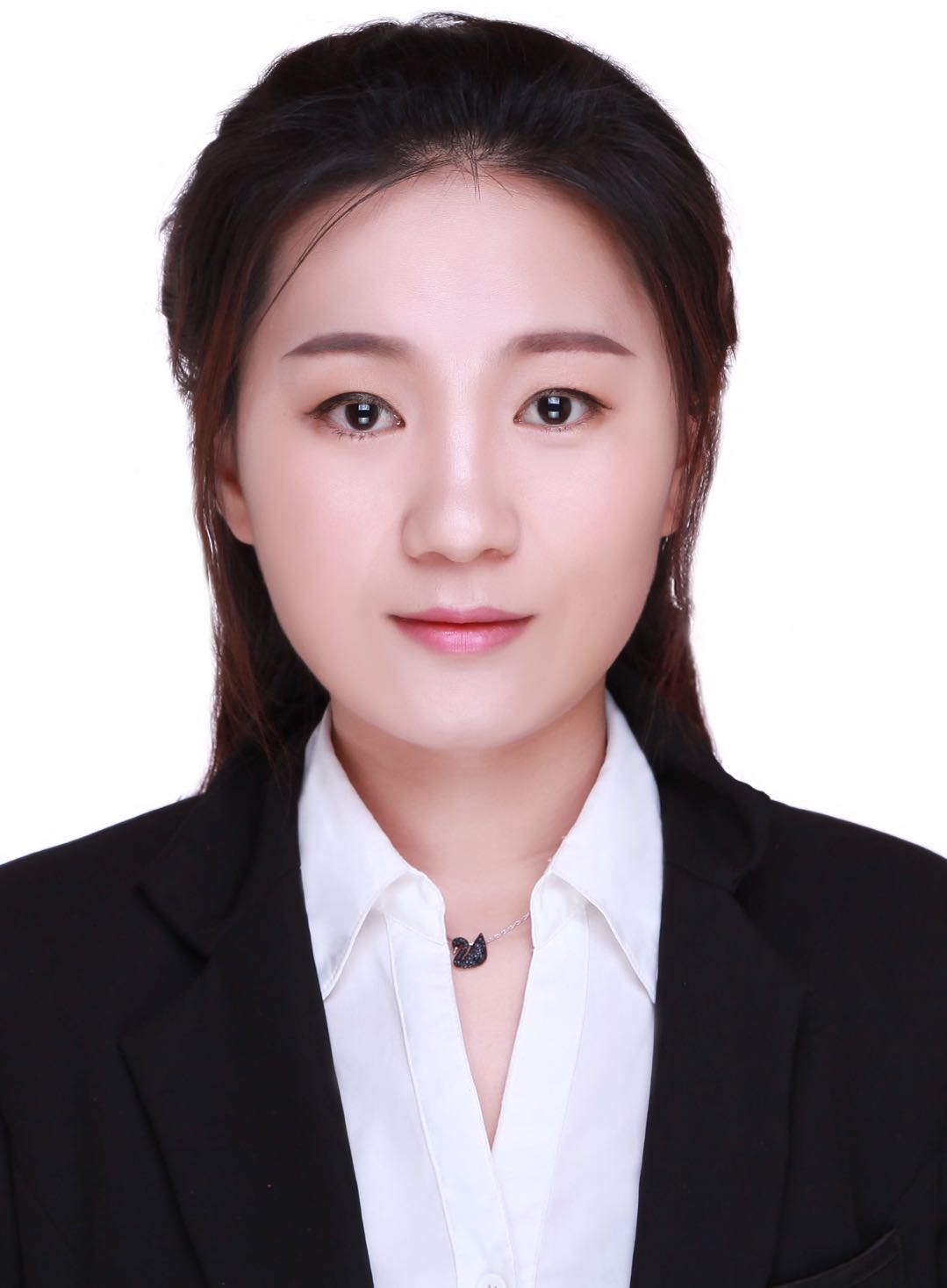}}]{Qian Li} is currently pursuing the Ph.D. degree with the Department of Computer Science and Engineering, Beihang University (BUAA), Beijing, China. Her research interests include text mining, representation learning, and event extraction.
\end{IEEEbiography}

 \vspace{-20pt}

\begin{IEEEbiography}[{\includegraphics[width=1in,height=1.25in,clip,keepaspectratio]{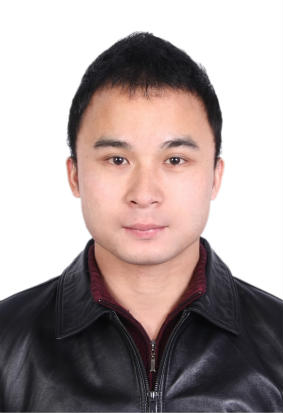}}]{Hao Peng} is currently an Assistant Professor at the School of Cyber Science and Technology, and Beijing Advanced Innovation Center for Big Data and Brain Computing in Beihang University. 
His research interests include representation learning, social network mining and reinforcement learning. 
To date, Dr Peng has published over 70 research papers in top-tier journals and conferences, including the IEEE TKDE, TC, TPDS, TITS, ACM TOIS, TKDD, TIST, TSAS, AAAI, IJCAI, SIGIR, Web Conference, CIKM, ICDM, DASFAA, etc.
\end{IEEEbiography}

 \vspace{-20pt}

\begin{IEEEbiography}[{\includegraphics[width=1in,height=1.25in,clip,keepaspectratio]{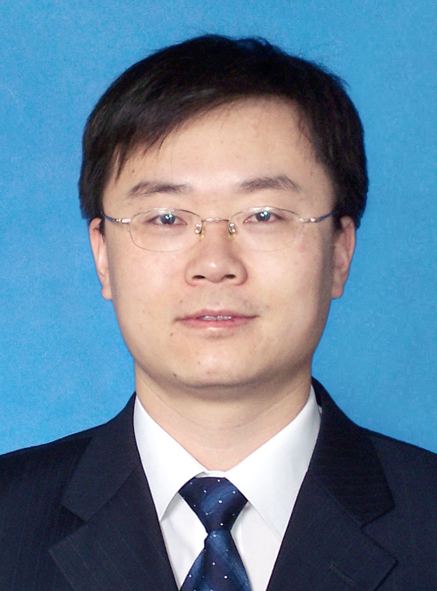}}]{Jianxin Li} is currently a Professor with the State Key Laboratory of Software Development Environment, and Beijing Advanced Innovation Center for Big Data and Brain Computing in Beihang University. His current research interests include social network, machine learning, big data and trustworthy computing.
\end{IEEEbiography}

 \vspace{-10pt}

\begin{IEEEbiography}[{\includegraphics[width=1in,height=1.25in,clip,keepaspectratio]{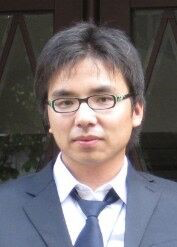}}]{Jia Wu} received the Ph.D. degree in computer science from the University of Technology Sydney, Ultimo, NSW, Australia. Dr Wu is currently an ARC DECRA Fellow in the Department of Computing, Macquarie University, Sydney, Australia. His current research interests include data mining and machine learning. Since 2009, he has published 100+ refereed journal and conference papers, including IEEE TPAMI, IEEE TKDE, IEEE TNNLS, IEEE TMM, ACM TKDD, NIPS, WWW, and ACM SIGKDD. Dr. Wu was the recipient of SDM’18 Best Paper Award in Data Science Track, IJCNN’17 Best Student Paper Award, and ICDM’14 Best Paper Candidate Award. He is the Associate Editor of the ACM Transactions on Knowledge Discovery from Data (TKDD) and Neural Networks (NN). He is a Senior Member of the IEEE.
\end{IEEEbiography}

 \vspace{-10pt}

\begin{IEEEbiography}[{\includegraphics[width=1in,height=1.25in,clip,keepaspectratio]{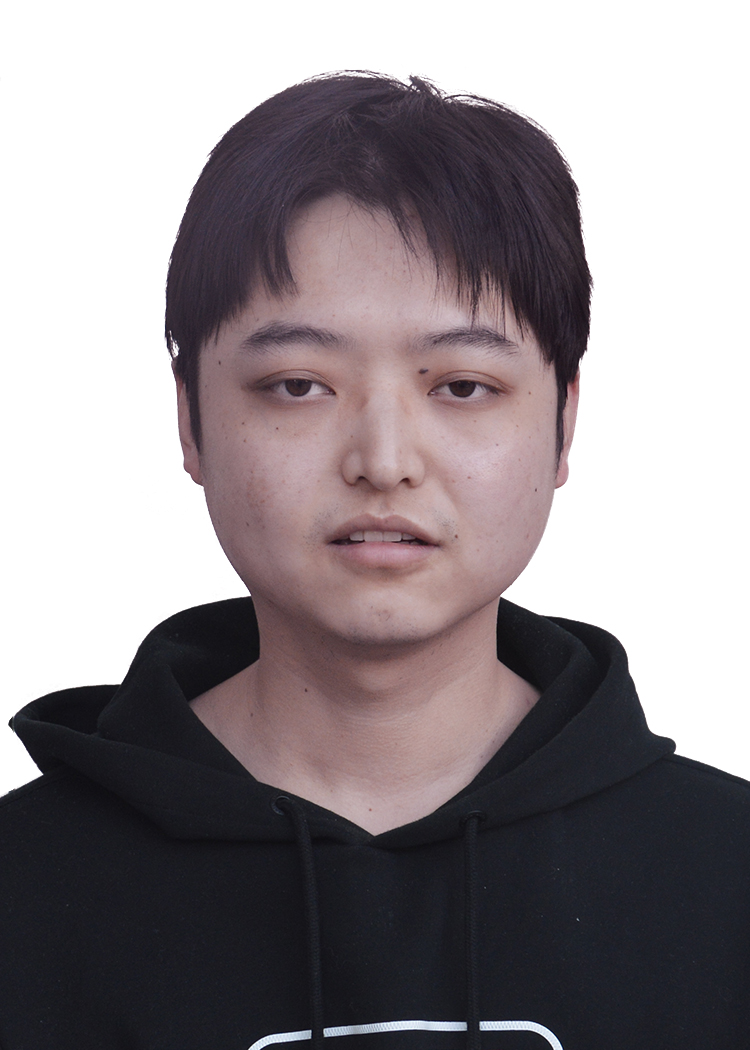}}]{Yanxing Ning} is currently pursuing the master degree with the Department of Computer Science and Engineering, Beihang University (BUAA), Beijing, China. Her research interests include text mining, representation learning, and event extraction.
\end{IEEEbiography}

 \vspace{-10pt}

\begin{IEEEbiography}[{\includegraphics[width=1in,height=1.25in,clip,keepaspectratio]{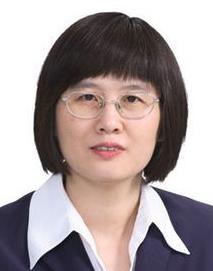}}]{Lihong Wang} is currently a Professor with the National Computer Network Emergency Response Technical Team/Coordination Center of China. Her current research interests include information security, cloud computing, big data mining and analytics, information retrieval, and data mining.
\end{IEEEbiography}
 \vspace{-10pt}

\begin{IEEEbiography}[{\includegraphics[width=1in,height=1.25in,clip,keepaspectratio]{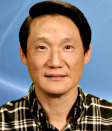}}]{Philip S. Yu} is a Distinguished Professor and the Wexler Chair in Information Technology at the Department of Computer Science, University of Illinois at Chicago, and also holds the Wexler Chair in Information Technology. Before joining UIC, he was at the IBM Watson Research Center, where he built a world-renowned data mining and database department. He is a Fellow of the ACM and IEEE. Dr. Yu was the Editor-in-Chiefs of ACM Transactions on Knowledge Discovery from Data (2011-2017) and IEEE Transactions on Knowledge and Data Engineering (2001-2004). He has received several IBM honors including 2 IBM Outstanding Innovation Awards, an Outstanding Technical Achievement Award, 2 Research Division Awards and the 94th plateau of Invention Achievement Awards. His research interest is in big data, including data mining, data stream, database and privacy. He has published more than 780 papers in refereed journals and conferences. He holds or has applied for more than 250 US patents.
\end{IEEEbiography}

 \vspace{-10pt}

\begin{IEEEbiography}[{\includegraphics[width=1in,height=1.25in,clip,keepaspectratio]{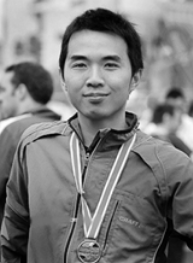}}]{Zheng Wang} is an Associate Professor with the University of Leeds, UK. His research cuts across the boundaries of parallel program optimization, systems security, and applied machine learning.
\end{IEEEbiography}

\end{document}